\documentclass{article}

\usepackage[main,preprint]{neurips_2026}

\usepackage[utf8]{inputenc} 
\usepackage[T1]{fontenc}    
\usepackage{hyperref}       
\usepackage{url}            
\usepackage{booktabs}       
\usepackage{amsfonts}       
\usepackage{nicefrac}       
\usepackage{microtype}      
\usepackage{xcolor}         

\usepackage{graphicx}
\usepackage{subcaption}

\usepackage{multirow}
\usepackage{wrapfig}
\usepackage{makecell}

\usepackage{algorithm}
\usepackage[noend]{algpseudocode}

\usepackage{mathtools}
\usepackage{amsfonts}
\usepackage{amsmath}
\usepackage{amsthm}
\usepackage{bm}
\usepackage{colonequals}
\usepackage{thmtools}

\usepackage[framemethod=TikZ]{mdframed}
\usetikzlibrary{matrix,decorations.pathreplacing}
\usetikzlibrary{decorations.pathreplacing,calligraphy}
\usepackage{tikzit}
\usetikzlibrary{positioning}

\tikzstyle{basic}=[fill=white, draw=black, shape=circle]
\tikzstyle{square}=[fill=white, draw=black, shape=rectangle]
\tikzstyle{big dashed}=[fill=white, draw=black, shape=circle, minimum width=1cm, dashed]
\tikzstyle{vertical ellipse dashed}=[fill=none, draw=blue, minimum width=0.75cm, minimum height=3cm, ellipse, dashed, tikzit shape=rectangle, tikzit draw=blue, tikzit fill=white]
\tikzstyle{small vertical ellipse dashed}=[fill=none, draw=blue, shape=circle, tikzit fill=white, tikzit draw=blue, dashed, minimum width=0.75cm, minimum height=1.5cm, tikzit shape=rectangle, ellipse]
\tikzstyle{tiny vertical ellipse dashed}=[fill=none, draw=blue, shape=circle, tikzit fill=white, ellipse, dashed, minimum width=0.75cm, minimum height=1cm, tikzit shape=rectangle]
\tikzstyle{red}=[fill=red, draw=black, shape=circle]
\tikzstyle{green}=[fill={rgb,255: red,0; green,128; blue,128}, draw=black, shape=circle]
\tikzstyle{blue}=[fill=blue, draw=black, shape=circle]
\tikzstyle{huge dashed}=[fill=white, draw=black, shape=circle, dashed, minimum width=2cm]
\tikzstyle{medium}=[fill=white, draw=black, shape=circle, minimum width=1cm]
\tikzstyle{pale green}=[fill={rgb,255: red,173; green,231; blue,0}, draw=black, shape=circle, minimum width=1cm]
\tikzstyle{horizontal ellipse dashed}=[fill=white, draw=black, tikzit draw=magenta, tikzit shape=rectangle, minimum width=3cm, minimum height=0.75cm, ellipse, dashed]
\tikzstyle{minsize}=[fill=white, draw=black, shape=circle, minimum width=0.75cm]
\tikzstyle{horizontal ellipse green}=[fill={rgb,255: red,191; green,255; blue,0}, draw=black, tikzit draw={rgb,255: red,191; green,255; blue,0}, tikzit shape=rectangle, minimum width=3cm, minimum height=0.75cm, ellipse, dashed]
\tikzstyle{horizontal ellipse blue}=[fill={rgb,255: red,107; green,203; blue,255}, draw=black, tikzit draw=blue, tikzit shape=rectangle, minimum width=3cm, minimum height=0.75cm, ellipse, dashed]
\tikzstyle{smallblack}=[fill=black, draw=black, shape=circle, inner sep=0 pt, minimum size=3 pt]
\tikzstyle{smallSquare}=[fill=white, draw=black, shape=rectangle, inner sep=0 pt, minimum size=6 pt]
\tikzstyle{smallCircle}=[fill=white, draw=black, shape=circle, inner sep=0 pt, minimum size=6 pt]
\tikzstyle{big vertical ellipse dashed}=[fill=none, draw=blue, shape=circle, tikzit shape=rectangle, ellipse, dashed, minimum width=0.95cm, minimum height=3.7cm]
\tikzstyle{smallred}=[fill=red, draw=red, shape=circle, inner sep=0 pt, minimum size=3 pt]
\tikzstyle{redfilled}=[fill={importantred!20}, draw=importantred, shape=circle, fill opacity=0.5]
\tikzstyle{bluefilled}=[fill={mathblue!20}, draw=mathblue, shape=circle, fill opacity=0.5]
\tikzstyle{greenfilled}=[fill={examplegreen!20}, draw=examplegreen, shape=circle]
\tikzstyle{orangefilled}=[fill={mathorange!20}, draw=mathorange, shape=circle, fill opacity=0.5]
\tikzstyle{new style 0}=[fill={rgb,255: red,191; green,191; blue,191}, draw=black, shape=circle]
\tikzstyle{small pink}=[fill={mathblue!20}, draw=mathblue, shape=circle, inner sep=0 pt, minimum size=6 pt, fill opacity=0.5]
\tikzstyle{smallorange}=[fill={mathorange!20}, draw=mathorange, shape=circle, inner sep=0 pt, minimum size=6 pt, fill opacity=0.5]
\tikzstyle{smallgreen}=[fill={examplegreen!20}, draw=examplegreen, shape=circle, inner sep=0 pt, minimum size=6 pt, fill opacity=0.5]

\tikzstyle{directed}=[->, -latex, draw=black]
\tikzstyle{undirected}=[-, draw=black]
\tikzstyle{directed red}=[draw=red, ->, -latex]
\tikzstyle{directed green}=[draw={rgb,255: red,0; green,128; blue,128}, ->, line width=1pt]
\tikzstyle{directed blue}=[draw=mathblue, ->, line width=1pt]
\tikzstyle{directed purple}=[draw={rgb,255: red,128; green,0; blue,128}, ->, line width=1pt]
\tikzstyle{undirected red}=[-, draw=red]
\tikzstyle{undirected green}=[-, draw={rgb,255: red,0; green,107; blue,61}, line width=1pt]
\tikzstyle{undirected blue}=[-, draw=blue]
\tikzstyle{undirected purple}=[-, draw={rgb,255: red,128; green,0; blue,128}, line width=1pt]
\tikzstyle{undirected dashed}=[-, line width=1pt, dashed, draw=black]
\tikzstyle{orange dashed}=[-, draw={rgb,255: red,255; green,128; blue,0}, dashed, line width=1.5pt]
\tikzstyle{directed dash}=[->, dashed, draw=black]
\tikzstyle{blue dashed}=[-, draw=blue, dashed, line width=1pt]
\tikzstyle{green dashed}=[-, draw={rgb,255: red,0; green,162; blue,0}, dashed, line width=1pt]
\tikzstyle{blue filled}=[-, fill={mathblue!20}, draw=mathblue, line width=1pt, tikzit fill=white]
\tikzstyle{red filled}=[-, fill={red!20}, line width=1pt, draw=red, opacity=0.5, tikzit fill=white]
\tikzstyle{green filled}=[-, draw=examplegreen, tikzit fill=white, fill={examplegreen!20}, line width=1pt]
\tikzstyle{orange filled}=[-, fill={mathorange!20}, draw=mathorange, line width=1pt, tikzit fill=white]
\tikzstyle{undirected dashed}=[-, draw={rgb,255: red,128; green,128; blue,128}, dashed, line width=1pt]
\tikzstyle{directed}=[->, -latex, fill=none, draw={rgb,255: red,128; green,128; blue,128}]
\tikzstyle{reddashed}=[-, draw=red, dashed]
\tikzstyle{blackdashed}=[-, dashed, draw=black]
\tikzstyle{black}=[-, ->, -latex, draw=black]
\tikzstyle{bluesolid}=[-, draw=mathblue]
\tikzstyle{bluedirected}=[-, ->, -latex, draw=mathblue]
\tikzstyle{pink dashed}=[-, dashed, draw=mathblue]
\tikzstyle{greensolid}=[-, draw=examplegreen]
\tikzstyle{greendirected}=[-, ->, -latex, draw=examplegreen]
\tikzstyle{green dashed}=[-, dashed, draw=examplegreen]
\tikzstyle{orangesolid}=[-, draw=mathorange]
\tikzstyle{orangedirected}=[-, ->, -latex, draw=mathorange]
\tikzstyle{orangedashed}=[-, dashed, draw=mathorange]

\tikzstyle{swr}=[fill=white, draw=black, shape=rectangle]
\tikzstyle{swc}=[fill=white, draw=black, shape=circle, minimum size=0.5 cm, inner sep=0.02cm]
\tikzstyle{mwc}=[fill=white, draw=black, shape=circle, minimum size=0.75cm, inner sep=0.02cm]
\tikzstyle{oval}=[fill=white, draw=black, shape=circle, minimum height=0.5cm, minimum width=1 cm, ellipse, inner sep=0.02cm]
\tikzstyle{mbc}=[fill={rgb,255: red,191; green,191; blue,191}, draw=black, shape=circle, minimum size=0.75cm, inner sep=0.02cm]
\tikzstyle{OvalRed}=[fill=white, draw=red, shape=circle, minimum height=0.5 cm, minimum width=1 cm, ellipse, inner sep=0.02 cm]
\tikzstyle{OvalBlue}=[fill=white, draw=blue, shape=circle, minimum height=0.5 cm, minimum width=1 cm, ellipse, inner sep=0.02 cm]
\tikzstyle{LongRectangle}=[fill=white, draw=black, shape=rectangle, minimum height=0.5cm, minimum width=1cm, inner sep=0.02cm]

\tikzstyle{sb}=[-]
\tikzstyle{new edge style 0}=[-, draw=red]

\usepackage{enumitem}

\theoremstyle{plain}
\newtheorem{theorem}{Theorem}[section]

\newtheorem{lemma}[theorem]{Lemma}

\theoremstyle{definition}
\newtheorem{definition}[theorem]{Definition}

\theoremstyle{remark}
\newtheorem{remark}[theorem]{Remark}

\usepackage{macros}
\title{Linear-time Rule Mining Under Formal Guarantees}

\author{%
  Jonathan Feldstein\thanks{Equal contribution.} \\
  Bennu.Ai\\
  \texttt{jonathan.feldstein@bennu.ai} \\
  \And
  Dominic Phillips\footnotemark[1] \\
  School of Informatics \\
  University of Edinburgh \\
  \texttt{dominic.phillips@ed.ac.uk} \\
  \AND
  Efthymia Tsamoura\thanks{This work was started before the author joined Huawei.} \\
  Huawei Labs \\
  Boole Research Center \\
  \texttt{efthymia.tsamoura@huawei.com} \\
}

\begin{document}

\maketitle

\begin{abstract}
Probabilistic logical models are a core component of neurosymbolic AI and are important in their own right for tasks that require high explainability. Unlike neural networks, logical theories that underlie the model are often handcrafted using domain expertise, making their development costly and prone to errors. While there are algorithms that learn logical theories from data, they are generally prohibitively expensive, limiting their applicability in real-world settings. Here, we introduce precision and recall for logical rules and define their composition as rule utility -- a measure of the predictive power of logical theories that can be computed in linear time in the data size. We then present SPECTRUM\footnote[1]{\url{https://github.com/jonathanfeldstein/SPECTRUM}}, a linear-time framework for learning logical theories from relational data, that mines recurrent patterns in the data, and subsequently uses our utility measure to evaluate and rank the rules derived from these patterns. Finally, we prove theoretical guarantees on the utility of the learnt logical theory. As a result, we demonstrate across various tasks that SPECTRUM scales to databases with orders of magnitude more relations and entities, often learning more accurate logical theories on CPUs in $< 1\%$ of the runtime of state-of-the-art neural network approaches on GPUs.
\end{abstract}
\section{Introduction}\label{section:intro}

\paragraph{Motivation.} \notion{Neurosymbolic AI} combines neural networks with (probabilistic) logical models to harness the strengths of both \citep{abstractNeSYSurvey, feldstein2024mapping}. 
These
frameworks outperform neural networks in several areas \citep{deepproblog, gu2019scene}, 
such as
interpretability \citep{mao2019neuro} and 
data efficiency \citep{concordia}. Unlike neural networks, which are trained from data, logical theories are typically handcrafted. 
This manual process is costly, prone to errors, and requires domain expertise.
As a result, there is a growing need to learn logical theories 
and probabilistic logical models (PLMs) from data -- a task known as \notion{structure learning}.

\paragraph{Limitations of SOTA.} 

The main difficulty of structure learning is the exponential cost of the problem w.r.t.
the length of 
rules and the number of relations in the data. Traditional techniques reduce the complexity of the problem by splitting the task into two steps: 
\notion{Pattern mining} - where frequently occurring substructures in the data are identified. \notion{Optimisation} - an iterative process to select the best logical formulae from a set of candidates identified using the mined patterns. However, 
frameworks to learn \notion{Markov logic networks} (MLNs) \citep{lsm, boostr, prism} and  \notion{probabilistic logical programs} \citep{mdie, diffILP, Inspire}  
still struggle
to scale beyond $\mathcal{O}(10^3)$ facts. To overcome the scalability issue, neural networks have been employed to learn logical rules \citep{DBLP:conf/nips/SadeghianADW19,rnnlogic, cheng2023neural}, scaling to $\mathcal{O}(10^5)$ facts. However, 
these frameworks require expensive 
hardware and still quickly reach memory limits \citep{cheng2023neural}.

\paragraph{Contributions.} We present SPECTRUM (\textbf{S}tructural \textbf{P}attern \textbf{E}xtraction and \textbf{C}heap \textbf{T}uning of \textbf{R}ules based on \textbf{U}tility \textbf{M}easure), an unsupervised model-agnostic (i.e. it can be used in conjunction with any PLM semantics, e.g. PSL, MLN, fuzzy logic) framework to learn rules.
Like traditional structure learners, SPECTRUM mines recurring patterns in the data. However, unlike prior approaches that rely on expensive \textit{exact} inference during rule optimisation, SPECTRUM uses an \textit{approximate} ranking of rules based on their joint quality. This allows SPECTRUM to scale linearly with dataset size and learn logical theories on datasets orders of magnitude larger than the prior art can handle.
Specifically, we make three contributions to tackle scalability, each linear in dataset size for a fixed maximum rule length, error tolerance, and predicate vocabulary: 

\begin{enumerate}[leftmargin=*]
    \item \notion{Linear-time rule evaluation}: We define \notion{utility} to quantify the predictive power of logical rules and theories, based on \notion{precision} (how often rules are satisfied) and \notion{recall} (how often rules predict data points). Utility is computed in linear time (w.r.t. number of relations) via subgraph counts. Utility enhances explainability by clarifying why rules are effective predictors; either through frequent correct predictions or coverage of large data portions.
    \item \notion{Linear-time pattern mining}: We present a linear-time algorithm (w.r.t. number of entities) that estimates the number of occurrences of subgraphs in the data and provide guarantees on the computational cost required for a given error in the corresponding utility estimates. 
    \item \notion{Linear-time optimisation}: We present a linear-time algorithm (w.r.t. number of relations) that 
    finds the best rules and sorts them by utility. 
    Since the utility measure is cheap, this effectively removes the optimisation step as a computational bottleneck in all but the very largest of theories. 
\end{enumerate}

\paragraph{Empirical results.} 
SPECTRUM scales easily to $\sim 100\times$ larger datasets than state-of-the-art (SOTA). On standard benchmarks, it matches or exceeds prior accuracy while running on CPUs in under $1\%$ of SOTA runtime on GPUs, effectively reducing energy consumption by $> 99.8\%$.


\section{Related Work} \label{section:related}

\paragraph{Inductive logic programming (ILP).} ILP has been used in diverse areas such as knowledge graph completion (KGC), program synthesis, robotics, explainable reinforcement learning, and visual scene understanding \citep{cropper2021inductive}. Here, we focus on ILP methods for learning relational theories over databases, and apply our framework to standard KGC benchmarks to evaluate it. Given a database, a set of positive (true) facts and a set of negative (false) facts, ILP techniques compute a theory that entails all positive and no negative facts.  
Approaches to learn Datalog theories are \notion{FOIL} \citep{FOIL}, \notion{MDIE} \citep{mdie} and \notion{Inspire} \citep{Inspire}, 
while approaches specifically designed for MLNs include \notion{BUSL} \citep{mihalkova_bottom-up_2007} and \notion{BOOSTR} \citep{boostr}. BOOSTR simultaneously learns the rules and their weights, using functional gradient boosting \citep{friedman2000greedy}. The inherent limitation of these approaches is the need for domain expertise to define patterns, as the quality of the theories, e.g. learnt by BOOSTR, drastically decreases in the absence of user-defined patterns \citep{prism}. \notion{LSM} \citep{lsm} learns MLNs 
by identifying patterns automatically by running random walks over a hypergraph representation of the data, but lacks guarantees on the quality of the mined patterns. \notion{PRISM} \citep{prism} 
mines patterns
with such guarantees, addressing this limitation but still failing to scale.

\paragraph{Differentiable methods.}
\notion{MetaAbd} \citep{meta-abd} mixes logical abduction (i.e. backward reasoning) \citep{Kakas2017} with ILP to simultaneously learn a logical theory and train a neural classifier. \notion{LRI} introduces meta-rules that soften the restrictions on templates \citep{campero_logical_2018}. \notion{HRI} \citep{pmlr-v162-glanois22a} further softens these to proto-rules. Recently, formulations of ILP where part of the computation can be differentiated (and backpropagated) have been proposed. For example, \notion{$\delta$ILP} uses fuzzy logic for interpreting rules \citep{diffILP}, and Sen et al. \cite{Sen_Carvalho_Riegel_Gray_2022} propose a technique based on \notion{logical neural networks} \citep{LNNs}.
However, these techniques still 
have scalability issues \citep{diffILP}.  
Several methods learn rules solely using neural networks, such as \notion{NeuralLP} \citep{NeuralLP}, \notion{DRUM} \citep{DBLP:conf/nips/SadeghianADW19}, \notion{NLMs} \citep{NLMs}, \notion{RNNLogic} \citep{rnnlogic}, and \notion{NCRL} \citep{cheng2023neural}. These techniques scale better but are limited by the shape of the rules that they learn: NeuralLP, DRUM, RNNLogic and NCRL learn chain rules  
while NLMs are restricted to rules where all head and body atoms contain the same variables.

\section{Preliminaries} \label{section:prelims}

\paragraph{First-order logic (FOL).} 
In FOL, \notion{constants} represent objects in the domain (e.g. $\constant{alice}$, $\constant{bob}$). \notion{Variables} range over the objects (e.g. $X$, $Y$, $Z$). Following standard practice in Datalog and the differentiable methods discussed in Section \ref{section:related}, we restrict ourselves to function-free FOL. In function-free FOL, a \notion{term} is a constant or a variable. 
An \notion{atom} has the form ${\pred(t_1,\dots,t_n)}$, where $\pred$ is an $n$-ary predicate relating the terms ${t_1,\dots,t_n}$.  
A Datalog rule $\Rule$, or simply \notion{rule}, is a FOL formula of the form
    $\forall \vec{X}.
    \bigwedge_{i=1}^n \pred_i(\vec{X}_i) \to 
    \pred(\vec{Y})$,
where $\bigwedge_{i=1}^n \pred_i(\vec{X}_i)$ is a \notion{conjunction} of atoms, $\to$ denotes logical \notion{implication},
$\vec{Y} \subseteq \bigcup_{i=1}^n\vec{X}_i$ and $\vec{X}_i \subseteq\vec{X}$. Quantifiers are commonly omitted.
The left and right sides of a rule are its \notion{body} $\body{\Rule}$ and
\notion{head} $\head{\Rule}$. The \notion{length} $L(\cdot)$ of a conjunction or rule  is the number of its atoms. 
A \notion{theory} $\Rules$ is a set of rules. In PLMs,  
each rule is associated with a weight representing the likelihood of the rule being satisfied. 
A \notion{ground} atom or \notion{fact} is the atom that results after replacing each occurrence of a variable with a constant.
\notion{Ground} conjunctions and rules are defined analogously.  A \emph{relational database} $\trainingdata$ is a set of facts. We denote a single atom as $\atom$, and its
ground counterpart -- a fact -- as $\logicfact$. Boldface denotes the
corresponding set: $\atoms$ is the set of all atoms, and $\groundatoms$ is the set of groundings of a particular $\atom$.


\begin{example}\label{example:fol}
        Consider recommender systems, where the goal is to predict whether a user will like an item based on user/item characteristics and prior user ratings. For a relational database $\trainingdata = \{\textsc{likes}(\mathtt{alice}, \mathtt{star wars}),$ $ \textsc{friends}(\mathtt{alice}, \mathtt{bob}), \textsc{likes}(\mathtt{bob}, \mathtt{star wars})\}$, the rule
    \begin{align*}
        \Rule_1: \textsc{friends}(U_1, U_2) \wedge \textsc{likes}(U_1, I) \rightarrow \textsc{likes}(U_2, I)
    \end{align*} 
    states that if two users are friends and one user liked an item, the other user will like it too.  
\end{example}

\begin{wrapfigure}[10]{R}{0.25\textwidth}
\centering
\begin{tikzpicture}
	\begin{pgfonlayer}{nodelayer}
		\node [style=mwc] (9) at (3, 0.25) {$\mathtt{s}$};
		\node [style=mwc] (10) at (1.95, -1) {$\mathtt{a}$};
		\node [style=mwc] (11) at (4.05, -1) {$\mathtt{b}$};
		\node [style=none, rotate=-50] (13) at (3.675, -0.25) {$\textsc{likes}$};
		\node [style=none, rotate=50] (14) at (2.325, -0.25) {$\textsc{likes}$};
		\node [style=none] (15) at (3, -0.875) {$\textsc{friends}$};
	\end{pgfonlayer}
	\begin{pgfonlayer}{edgelayer}
		\draw [style=undirected] (9) to (10);
		\draw [style=undirected] (9) to (11);
		\draw [style=undirected red] (10) to (11);
	\end{pgfonlayer}
\end{tikzpicture}
        \caption{Datagraph $\datagraph$ for $\trainingdata$ in Example \ref{example:fol}.}\label{fig:data}
\end{wrapfigure}

\paragraph{Hypergraphs.} A \notion{hypergraph} ${\graph}$ is a pair $(V,E)$, where $V$ is a set of \notion{nodes} and $E$ is a set of \notion{edges} where each element of $E$ is a set of nodes $\{v_1, \dots, v_n\}$ from $V$. For brevity, we refer to hypergraphs simply as graphs. A graph ${\graph}$ is \notion{labelled} if each edge $e$ in ${\graph}$ is labelled with a categorical value. A \notion{path} is an alternating sequence of nodes and edges, $(v_{1}, e_{1}, \dots, e_{l}, v_{l+1})$, where each edge $e_i$ contains $v_i$ and $v_{i+1}$. 
${\graph}$ is \notion{connected} if there exists a path between any two nodes.  
A relational database $\trainingdata$ can be represented by a graph $\datagraph=(V,E)$, 
where, for each constant $\constant{c}_i$ in $\trainingdata$, $V$ includes a node,
and, for each fact ${\pred(\constant{c}_1,\dots, \constant{c}_{k})}$ in $\trainingdata$, $E$ includes an edge ${e=}{\{\constant{c}_1,\dots, \constant{c}_{k}\}}$ 
with label $\pred$. Fig.~\ref{fig:data} illustrates the datagraph for Example \ref{example:fol}.

A notation table for all quantities used in this paper is in App.~\ref{app:notation}.


\section{Rule Utility} \label{section:utility}

This section introduces 
\notion{utility} -- a measure to assess the ``usefulness" of a theory without requiring inference. 
App.~\ref{app:design_choices} explains the design choices made in this section, and compares with similar metrics 
in the literature.
The definitions that follow rely on counts of 
recurring subgraphs in a datagraph, which we call 
\notion{patterns}.


\begin{wrapfigure}[16]{R}{0.25\textwidth}
\centering
\begin{tikzpicture}
	\begin{pgfonlayer}{nodelayer}
		\node [style=mwc] (0) at (0, 0.25) {$I$};
		\node [style=mwc] (1) at (-1, -1) {$U_1$};
		\node [style=mwc] (2) at (1, -1) {$U_2$};
		\node [style=none] (5) at (0, -1.65) {$\pattern{\body{\Rule_1} \wedge \head{\Rule_1}}$};
		\node [style=none,rotate=-50] (6) at (0.65, -0.25) {$\textsc{likes}$};
		\node [style=none, rotate=50] (7) at (-0.65, -0.25) {$\textsc{likes}$};
		\node [style=none] (8) at (0, -0.85) {$\textsc{friends}$};
		\node [style=mwc] (16) at (0, 2.75) {$I$};
		\node [style=mwc] (17) at (-1, 1.5) {$U_1$};
		\node [style=mwc] (18) at (1, 1.5) {$U_2$};
		\node [style=none] (19) at (0, 1) {$\pattern{\body{\Rule_1}}$};
		\node [style=none, rotate=50] (21) at (-0.65, 2.25) {$\textsc{likes}$};
		\node [style=none] (22) at (0, 1.65) {$\textsc{friends}$};
	\end{pgfonlayer}
	\begin{pgfonlayer}{edgelayer}
		\draw [style=undirected] (0) to (1);
		\draw [style=undirected] (0) to (2);
		\draw [style=undirected red] (1) to (2);
		\draw [style=undirected] (16) to (17);
		\draw [style=undirected red] (17) to (18);
	\end{pgfonlayer}
\end{tikzpicture}
        \caption{The patterns of  $\Rule_1$ and $\body{\Rule_1}$. }\label{fig:patterns}
\end{wrapfigure}

\paragraph{Patterns.} 

Conjunctions of atoms can be viewed as graphs. For a conjunction $\logicformula \vcentcolon= \bigwedge_{i=1}^n \pred_i(\vec{t}_i)$, the \textit{pattern} of $\logicformula$, denoted $\pattern{\logicformula}=(V,E)$, is the graph where, for each term~$t_i$ occurring in $\logicformula$, $V$ includes a node $t_i$, and, for each atom $\pred(t_1,\dots,t_n)$ occurring in $\logicformula$, $E$ includes an edge 
${\{t_1,\dots,t_n\}}$ with label $\pred$. Given a rule $\Rule$, the patterns corresponding to its head and body are denoted by $\pattern{\head{\Rule}}$ and $\pattern{\body{\Rule}}$. We call $\pattern{\body{\Rule}\wedge \head{\Rule}}$ the \notion{rule pattern} of $\Rule$. Fig.~\ref{fig:patterns} shows the (rule) patterns of $\Rule_1$ and $\body{\Rule_1}$ from Example~\ref{example:fol}.
Rule $\Rule$ is \notion{connected} if $\pattern{\body{\Rule}\wedge \head{\Rule}}$ is connected; it is \notion{body-connected} if $\pattern{\body{\Rule}}$ is connected.
%
A \notion{ground pattern}  of a conjunction $\logicformula$ is
the graph corresponding to a grounding of $\logicformula$ that is satisfied in $\trainingdata$. 
We denote by $\instances{\body{\Rule}\wedge \head{\Rule}}$ the set of all 
ground patterns of $\body{\Rule} \wedge \head{\Rule}$ in $\trainingdata$. 
For a fact $\logicfact$ that is a grounding of $\atom = \head{\Rule}$, 
we use $\restrictedinstances{\logicfact}$ to denote the subset of $\instances{\body{\Rule}\wedge \head{\Rule}}$ containing only groundings of patterns of the form $\pattern{\body{\Rule}\wedge \logicfact}$. A rule is \notion{term-constrained} if every term occurs in at least two atoms, e.g. $\Rule_1$ in Example~\ref{example:fol} is term-constrained but $\textsc{friends}(U_1, U_2) \rightarrow \textsc{likes}(U_2, I)$ is not, since neither $U_1$ nor $I$ appears twice.

The following definitions hold for connected and body-connected rules:

\begin{definition} \label{definition:precision}
    The \textbf{precision} of rule $\Rule$ is defined as 
    \begin{align*}
            \precision{\Rule} \colonequals \frac{\Count{\instances{\body{\Rule}\wedge \head{\Rule}}}}{\Count{\instances{\body{\Rule}}}}. 
    \end{align*}
\end{definition}

Intuitively, $\precision{\Rule}$ is the fraction of times that the head and body of a rule are both true when the body is true in the data. If one considers cases where the body and head are true as true positives TP (the rule is satisfied), and cases where the body is true but the head is false as false positives FP (the rule is not satisfied), then this definition is analogous to the definition of precision in classification tasks: $\precision{\Rule} = TP / (TP + FP)$. Useful rules should 
often be true,
and thus have high precision. 

Precision, as defined above, underestimates how often a rule is satisfied if the rule has symmetries. We fix this issue
by multiplying the precision by a symmetry factor:

\begin{definition}
\label{definition:symmetry}
    The \textbf{symmetry factor} of rule $\Rule$, denoted by $\symmetry{\Rule}$, is defined as the number of subgraphs in $\pattern{\body{\Rule} \wedge \head{\Rule}}$ that are isomorphic to $\pattern{\body{\Rule}}$. 
\end{definition}

\begin{example}\label{example:symmetry}
    The database $\trainingdata$ in Example \ref{example:fol} has two ground 
    patterns of $\body{\Rule_1}$ (the top graph in Fig.~\ref{fig:patterns} can be found twice in Fig.~\ref{fig:data}), and one ground pattern of $\body{\Rule_1} \wedge \head{\Rule_1}$ (the bottom graph in Fig.~\ref{fig:patterns} can be found once in Fig.~\ref{fig:data}).  Hence, $\precision{\Rule_1} = \frac{1}{2}$ even though the rule is always satisfied. However, rule $\Rule_1$ has a symmetry factor $\symmetry{\Rule_1}=2$ (the top graph in Fig.~\ref{fig:patterns} can be found twice in the bottom graph). Thus, $\precision{\Rule_1}\cdot \symmetry{\Rule_1}=1$, as expected since the rule is always satisfied.
\end{example}

Further, if the facts in $\trainingdata$ are unbalanced (i.e. facts of different predicates occur with different frequencies), then certain rules can still have high precision even if the facts are uncorrelated. 
We fix this issue
by dividing the precision by a Bayesian prior:

\begin{definition}
\label{definition:prior}
    The \textbf{Bayesian prior} of rule~$\Rule$ is defined as
    \begin{align*}
        \prior{\Rule} := \frac{\Count{\instances{\head{\Rule}}}}{ \sum_{\atom \in \mathcal{A}} \Count{\instances{\atom}}},
    \end{align*}
    where $\mathcal{A}$ is the set of all atoms constructible over all predicates in $\trainingdata$ with the same tuple of terms as $\head{\Rule}$.
\end{definition}

The Bayesian prior can be thought of as the prior probability, given a fixed body, of observing that predicate in the head. A useful rule should have a symmetry-corrected precision, $\precision{\Rule} \cdot \symmetry{\Rule}$, that is better than prior random chance, $\prior{\Rule}$, i.e. $\frac{\precision{\Rule} \cdot \symmetry{\Rule}}{\prior{\Rule}} > 1$.

\begin{example}
\label{exa:bayes}
Consider, in addition to rule $\Rule_1$, rule $\Rule_2$:
\begin{align*}
    \Rule_2 : \textsc{likes}(U_1, I) \wedge \textsc{friends}(U_1, U_2) \rightarrow \textsc{dislikes}(U_2, I)
\end{align*}
Assume also that the facts in $\trainingdata$ follow these statistics: 
(i) if a user $U_1$ likes an item $I$, then a friend of theirs $U_2$ likes $I$ 
with probability 50\% and dislikes $I$ with probability 50\%, and
(ii) the number of $\textsc{dislikes}$-facts is $10 \times$ larger than the number of 
$\textsc{likes}$-facts.
From assumption (i) and Def.~\ref{definition:precision}, it follows that $\precision{\Rule_1}=\precision{\Rule_2}=\frac{1}{2}$ since 
for each grounding of the body 
${\textsc{likes}(\constant{u_1}, \constant{i}) \wedge \textsc{friends}(\constant{u_1}, \constant{u_2})}$ in $\trainingdata$,  
there is either a fact ${\textsc{likes}(\constant{u_2}, \constant{i}) }$ or a fact 
${\textsc{dislikes}(\constant{u_2}, \constant{i}) }$ in $\trainingdata$ and each fact has a probability of 50\%. If there were no correlations in the facts (i.e. assumption (i) did not hold), then we would expect, from assumption (ii), that the head is $10 \times$ more likely to be \textsc{dislikes} than \textsc{likes}. The result $\precision{\Rule_1} = \precision{\Rule_2} = \frac{1}{2}$ is, therefore, misleading, since $\Rule_1$ is correct more than five times as often as random chance ($1/2$ vs $1/11$) and $\Rule_2$ is correct less often than random chance ($1/2$ vs $10/11$). 

We now compute the Bayesian priors.
In both cases, the head predicate contains a user and an item term, thus, $\mathcal{A} = \{\textsc{likes}(U,I), \textsc{dislikes}(U,I)\}$ 
in Def.~\ref{definition:prior}. 
Writing $n_{\textsc{l}} = |\instances{\textsc{likes}(U,I)}|$ and $n_{\textsc{d}} = |\instances{\textsc{dislikes}(U,I)}|$, the Bayesian priors are thus:
\[
  \prior{\Rule_1} = \frac{n_{\textsc{l}}}{n_{\textsc{l}} + n_{\textsc{d}}} = \frac{1}{10+1}, \quad
  \prior{\Rule_2} = \frac{n_{\textsc{d}}}{n_{\textsc{l}} + n_{\textsc{d}}} = \frac{10}{10+1}.
\]
    Notice that these are exactly the probabilities of $\Rule_1$ and $\Rule_2$ being true if the facts in $\trainingdata$ were uncorrelated. Precision of the rules, 
    corrected for this prior, would 
    then be $\frac{\precision{\Rule_1}}{\prior{\Rule_1}} = \frac{11}{2}$ and $\frac{\precision{\Rule_2}}{\prior{\Rule_2}} = \frac{11}{20}$. Since $\frac{\precision{\Rule_1}}{\prior{\Rule_1}} >1$, $\Rule_1$ successfully predicts a correlation. In contrast, since $\frac{\precision{\Rule_2}}{\prior{\Rule_2}} < 1$, $\Rule_2$
    makes a prediction that is worse than a random guess. 
   Hence, 
   $\Rule_1$ is correctly identified as more useful than $\Rule_2$.
\end{example}

In addition to being precise, useful rules should account for many diverse observations. We introduce a metric to count how often a rule pattern recalls facts in the data.

\begin{definition}\label{definition:log-recall-rule}
    The \textbf{recall} of rule $\Rule$ is defined as
    \begin{align*}
        \recall{\Rule} \colonequals \sum_{\logicfact \in \groundatoms} \ln\left(1 + \left|\restrictedinstances{\logicfact}\right|\right),
    \end{align*}
    where $\groundatoms$ is the set of all groundings of $\atom \colonequals \head{\Rule}$ in $\trainingdata$.
\end{definition}

Intuitively, $\left|\restrictedinstances{\logicfact}\right|$ 
says how many different {groundings} of $\Rule$ entail $\logicfact$, and the logarithm reflects the diminishing returns of recalling the same fact. {Importantly,} recalling the same fact increases {recall} logarithmically, {while} recalling different facts increases {recall} linearly.

Longer rules are biased to have more groundings in the data than shorter rules, due to a combinatorial explosion of possible groundings. Therefore, recall is biased towards longer rules. 
To address this, we introduce a complexity factor that penalises longer rules: 

\begin{definition}
\label{definition:complexity}
The \textbf{complexity factor} of $\Rule$ of length $L(\Rule)$ is defined as
\begin{align*}
    {\complexity{\Rule} \colonequals e^{-L(\Rule)}}.
\end{align*}
\end{definition}

The complexity-corrected recall, $\recall{\Rule}\cdot\complexity{\Rule}$, discourages using longer rules if they do not have a correspondingly larger recall, thus favouring the simplest explanation for the data (Occam's razor).

Useful rules should exhibit both high precision (corrected for symmetry and priors) and high recall (corrected for complexity). This leads to a natural metric for quantifying the ``usefulness" of a rule:

\begin{definition}\label{definition:utilityRule} 
    The \textbf{utility of rule} $\Rule$ is defined as
    \begin{align*}
       \utility{\Rule} \colonequals \frac{\precision{\Rule} \symmetry{\Rule}}{\prior{\Rule}} \cdot \recall{\Rule}\complexity{\Rule}. 
    \end{align*}
\end{definition}

Finally, we extend the notion of utility to a theory $\Rules$. 
Different rules can recall the same fact. The recall for a set of rules should be analogous to Def.~\ref{definition:log-recall-rule} but include contributions from all rules:

\begin{definition} \label{definition:rule-set-recall} 
For a set of rules $\Rules_\alpha$ with the same head~$\alpha$, the \textbf{complexity-corrected rule-set recall} is defined as $\recall{\Rules_\alpha} \cdot \complexity{\Rules_\alpha}$, where 
\[
\recall{\Rules_\atom} \colonequals \sum_{\logicfact \in \groundatoms}\ln \left(1  + \sum_{\Rule \in \Rules_\atom} \left|\restrictedinstances{\logicfact}\right| \right)
\quad \text{and} \quad
\complexity{\Rules_\atom} \colonequals \left(\prod_{\Rule \in \Rules_\atom} \complexity{\Rule}\right)^{(|\Rules_\alpha|)^{-1}}.
\]
\end{definition}

Intuitively, $\sum_{\Rule \in \Rules_\atom} \left|\restrictedinstances{\logicfact}\right|$ counts the number of different instantiations of \emph{all} rules in the set $\Rules_\atom$ that entail a particular fact $\logicfact$, whereas $\complexity{\Rules_\atom}$ is simply the geometric mean of the complexity factors of all rules in set $\Rules_\atom$.
We are now ready to introduce the notion of theory utility:

\begin{definition}
\label{definition:utilityRuleSet}
    The \textbf{utility of theory} $\Rules$ is 
    \begin{align*}
        {\utility{\Rules} \colonequals 
        \sum_{\atom \in \atoms}\left(\sum_{\Rule \in \Rules_{\atom}} \frac{\precision{\Rule} \symmetry{\Rule}}{\prior{\Rule}}\right) \cdot  \recall{\Rules_{\atom}} \complexity{\Rules_{\atom}}},
    \end{align*}
    where $\atoms = \{ \head{\Rule}  \, | \, 
     \Rule \in \Rules\}$ and  $\Rules_\atom = \{\Rule \in \Rules \, | \, \head{\Rule} = \alpha \}$.
\end{definition}

The outer sum in Def.~\ref{definition:utilityRuleSet} runs over the different atoms 
occurring in the heads of the rules in $\Rules$, and the inner sum runs over the different rules with the same head atom. Computing rule utility requires enumerating all 
ground patterns of a rule, its body, and its head in the data. In the worst case, the complexity of computing the theory utility is $\mathcal{O}(|\Rules|(|E|+D+|\mathcal{P}|))$, where $|E|$ is the number of relations in the dataset, $D$ is the maximum rule length and $|\mathcal{P}|$ is the number of distinct predicates in the dataset (proof in App.~\ref{appx:utilityComplexity}). 

\section{SPECTRUM}
\label{sec:SPECTRUM}

\paragraph{Objective.} Given a datagraph $\datagraph= (V,E)$, find the theory $\Rules$ consisting of up to $M$ rules of length up to $L_{\text{max}}$ that collectively maximise the theory utility $\utility{\Rules}$.

\paragraph{Approach.} SPECTRUM tackles this objective by: 
\begin{enumerate}[leftmargin=*]
    \item Mining up to $N|V|$ patterns of each length $l\in \{1, \dots, L_{\text{max}}\}$ (Section~\ref{section:pattern mining}),
    \item Generating candidate rules that can be instantiated from the mined patterns (Section~\ref{section:optimization}),
    \item Creating a theory of (up to) $M$ rules from the set of candidate rules by approximately maximising rule-set utility (Section~\ref{section:optimization}).
\end{enumerate}

In Section~\ref{section:theory}, we present completeness and noise robustness guarantees for SPECTRUM, and show how to automatically set $N$ to achieve rule-utility estimates with a 
(maximal) fractional uncertainty $\varepsilon$.

SPECTRUM thus has three hyperparameters: $M$, the maximum number of rules in the final theory; $D$, the mining depth (which sets the maximum rule length $L_{\text{max}} = 2D + 1$); and $\varepsilon$, the target uncertainty, balancing a trade-off between accuracy and computational cost.

\paragraph{Shape of the Mined Rules.} Every structure learning framework imposes  
restrictions on the shape of the mined rules.
In particular, all recent neural methods mine \notion{chain} rules over \emph{binary} predicates \citep{NeuralLP,DBLP:conf/nips/SadeghianADW19,lajus_fast_2020,rnnlogic,cheng2023neural}.
Similarly, a rule mined by SPECTRUM is
\begin{enumerate}[leftmargin=*]
    \item[(1)] body-connected, term-constrained, and
    \item[(2)] \emph{either} contains only unary predicates \emph{or} each of its terms occurs in at most one unary and two binary predicates.
\end{enumerate}
In (1), term-constrainedness ensures each term is in at least one known atom, aiding link prediction, while body-connectedness is required for computing utility (cf.\ Section~\ref{section:utility}). Restriction (2) is simply an artefact of the specific pattern mining algorithm we present in Section~\ref{section:pattern mining}.\footnote[2]{The pattern mining component may be replaced by a more sophisticated algorithm, e.g.\ mining patterns with higher-arity predicates or trees in addition to cycles, without affecting the rest of the pipeline. We leave this for future work.} If the data contains only binary predicates, these restrictions reduce to the chain rules imposed by prior art \citep{cheng2023neural,lajus_fast_2020,rnnlogic}. However, by allowing unary predicates, we are able to model non-chain rules that are important in real-world datasets (examples in App.~\ref{sec:appx:KGC}). Nevertheless, the shapes of rules learnt by SPECTRUM are still more restricted than those of general ILP methods \cite{law2020ilasp,cropper2021learning} or template-based rule mining \citep{cropper2016metagol,pmlr-v162-glanois22a,muggleton1991inductive}. In Appendix~\ref{appx:ExpressivityOfRuleShapes} we contrast the rule shapes learnable by SPECTRUM with related work in more detail, and explain the trade-off with respect to scalability. 

\subsection{Pattern Mining} \label{section:pattern mining}

This section presents our technique for mining patterns, i.e.
subgraphs in a relational database graph. Finding all
subgraphs is generally a hard problem with no known polynomial algorithm \citep{bomze_maximum_1999}. Prior work includes pruning techniques and canonical ordering to ameliorate the cost, but without changing the complexity \citep{alam2021mining}, or mines only maximal cliques, in $\mathcal{O}(|V|^{2.376})$ time \citep{spyropoulou2014interesting}. In contrast, we present a non-exhaustive algorithm that has only $\mathcal{O}(|V|)$ complexity
in the dataset size and mines more general patterns (not just cliques) up to a given size. We show that our approach allows us to compute estimates of
utility that are \textit{close}, in a precise sense, to their true values.

Intuitively, Alg.~\ref{alg:fragmentMining} searches for ground patterns by running paths in parallel from a starting node. This is done by calling a recursive function \alg{NextStep} from each node $v_0$ in $\datagraph$ (lines \ref{line:main1}-\ref{line:main2}). The recursion stops at a \notion{maximum depth} $D$ and a \notion{maximum number of paths}~$N$, which SPECTRUM sets from $\varepsilon$ (Section~\ref{section:theory}). We use $\restrictedunaries{v}{\graph}$ and $\restrictedbinaries{v}{\graph}$ to denote the sets of unary and binary edges in the graph $\graph$ that are incident to $v$. A step-by-step example of Alg.~\ref{alg:fragmentMining} can be found in App.~\ref{section:pattern_mining_example}.

\begin{algorithm}[htb]
\caption{\alg{MinePatterns}($\datagraph$, $D$, $N$)}
\label{alg:fragmentMining}

\begin{algorithmic}[1]

\For{\textbf{each} $v_0 \in \datagraph$} \label{line:main1}
    \State $\globalpatterns \gets \globalpatterns \cup$ \Call{GetUnaryPatterns}{$v_0$} \label{line:doubleunary}
    \State \Call{NextStep}{$\datagraph, v_0, D, N, 0, \{\emptyset\}, \emptyset$} \label{line:main2}
\EndFor
\State \Return $\globalpatterns$
\vspace{0.1cm}

\Function{NextStep}{$\datagraph, v, D, n, d, \currentpatterns, \previousedges$}
    \State $\newpatterns \gets \emptyset$
    \For{$\underline{\mathcal{G}} \in \currentpatterns$} \label{line:unarygrafting1}
        \For{$e \in \restrictedunaries{v}{\datagraph}$} \Comment{Unary grafting}
            \State $\newpatterns \gets \newpatterns \cup \{\underline{\mathcal{G}} \circ e\}$ \label{line:unarygrafting2}
        \EndFor
    \EndFor
    \State $\globalpatterns \gets \globalpatterns \cup \newpatterns$ \label{line:unarystoring}

    \If{$d < D$} \label{line:maxrecursion}
        \State $\edges' \gets \restrictedbinaries{v}{\datagraph} \setminus \previousedges$ \label{line:binaryselection1}
        \If{$n < |\edges'|$} \label{line:Ncheck}
            \State $\edges' \gets$ \Call{SelectRandomElements}{$\edges', n$} \label{line:selection}
            \State $n' \gets 1$ \label{line:N1}
        \Else
            \State $n' \gets \lceil n / |\edges'| \rceil$ \label{line:binaryselection2}
        \EndIf

        \For{$e = \{v, v'\} \in \edges'$} \label{line:binarygrafting1}
            \State $\finalpatterns \gets \emptyset$
            \For{$\underline{\mathcal{G}} \in \newpatterns$} \Comment{Binary grafting}
                \State $\finalpatterns \gets \finalpatterns \cup \{\underline{\mathcal{G}} \circ e\}$ \label{line:binarygrafting2}
            \EndFor
            \State $\globalpatterns \gets \globalpatterns \cup \finalpatterns$ \label{line:binarystoring}
            \State $\previousedges \gets \previousedges \cup \{e\}$
            \State \Call{NextStep}{$\datagraph, v', D, n', d~+~1, \finalpatterns, \previousedges$} \label{line:nextcall}
        \EndFor
    \EndIf
\EndFunction
\end{algorithmic}
\end{algorithm}

In each call of \alg{NextStep}, the algorithm visits a node $v \in V$ of $\datagraph$. 
At node $v$, each unary edge in $\restrictedunaries{v}{\datagraph}$ is grafted onto each previously found ground pattern $\underline{\mathcal{G}} \in \currentpatterns$ (line~\ref{line:unarygrafting1}-\ref{line:unarygrafting2}). 
We use $\underline{\mathcal{G}} \circ e$ to denote the ground pattern that results after adding edge~$e$ and the nodes of $e$ to $\underline{\mathcal{G}}$. The resulting ground patterns $\unarypatterns$ are stored in $\globalpatterns$ (line~\ref{line:unarystoring}).
If the maximum mining depth has not been reached (line~\ref{line:maxrecursion}), a subset of the binary edges of node $v$ is then selected (lines \ref{line:binaryselection1}-\ref{line:binaryselection2}). 
The algorithm avoids mining  patterns corresponding to tautologies by excluding previously visited binary edges (line~\ref{line:binaryselection1}). 
To keep the complexity linear, we limit the maximum number of paths to  $N$ by setting the maximum number of selected binary edges $n$ to be $N$ divided by the number of binary edges selected at each previous stage (lines~\ref{line:selection}, \ref{line:binaryselection2}). 
We graft each chosen binary edge onto the ground patterns in $\newpatterns$ (lines~\ref{line:binarygrafting1}-\ref{line:binarygrafting2}), store the resulting ground patterns $\finalpatterns$ in $\globalpatterns$ (line~\ref{line:binarystoring}), and pass $\finalpatterns$ on to the next recursive call to extend these patterns further (line~\ref{line:nextcall}). 
In the next recursive call, the current mining depth $d$ is increased by $1$, expanding the search of ground patterns to include nodes up to a distance $d$ away from $v_0$.


At each depth $d \in \{0, \dots, D-1\}$ the algorithm grafts up to one unary edge and one binary edge onto the patterns previously discovered along that path. At depth $d=D$, the algorithm only grafts up to one unary edge onto the patterns, since it terminates before grafting binaries (line~\ref{line:maxrecursion}). Thus, the patterns found by Alg.~\ref{alg:fragmentMining} are of maximum length $2D+1$.

As a special case, Alg.~\ref{alg:fragmentMining}
mines patterns consisting of multiple unary edges on a single node (line~\ref{line:doubleunary}) to
find rules of the form ${\pred_1(X) \wedge \dots \wedge \pred_{m-1}(X) \rightarrow \pred_m(X)}$. For the $n$ unary edges incident to a node, $\alg{GetUnaryPatterns}$ generates all ${{n}\choose{m}}$ subsets of these predicates of size $m \in \{1, \dots, n\}$.

\subsection{Rule Optimisation}\label{section:optimization}


Given a map $\globalpatterns$ of mined patterns to their grounding, \alg{OptimiseRules} aims to find the best subset of rules to explain the data. Each pattern $\pattern{\varphi}$ in the keys of the map $\globalpatterns$ is considered in turn. Each rule that could have resulted in this pattern, i.e. a rule from the set $\mathcal{R} \vcentcolon= {\{\Rule\mid \pattern{\body{\Rule} \wedge \head{\Rule}} = \pattern{\varphi}\}}$, is considered. If a rule $\Rule \in \mathcal{R}$ is term-constrained and satisfies $\frac{\precision{\Rule} \symmetry{\Rule}}{\prior{\Rule}} > 1$, i.e. the rule is a better predictor than a random guess (Section~\ref{section:utility}), then $\Rule$ is added to the set of candidate rules $\Rules_{\textnormal{candidates}}$.


From the set of candidate rules, a subset of $M$ rules with the highest individual utility is chosen (Def.~\ref{definition:utilityRule}). The utility of each rule $\Rule$ is directly calculated from $\globalpatterns$; since this is a map from patterns $\pattern{\varphi}$ of a conjunction $\varphi$ to its groundings in $\datagraph$, i.e. $\instances{\varphi}$. The quantity $\Count{\instances{\varphi}}$ can be looked up in the map. The conjunction $\varphi$ can be $\body{\Rule}$, $\head{\Rule}$, or $\body{\Rule}\wedge\head{\Rule}$. The complexity and symmetry factor can be computed for each $\Rule$ directly from its length, rule pattern and body pattern.


SPECTRUM then orders the remaining $M$ rules 
by their contribution to the theory utility (Def.~\ref{definition:utilityRuleSet}). 
The algorithm starts by finding the rule with the highest utility and stores it in a vector $\Rules_{\textnormal{final}}$. Then, in each iteration of the while-loop, SPECTRUM finds the rule out of the remaining ones that provides the highest increase in theory utility when added to the current rules in $\Rules_{\textnormal{final}}$. 


\begin{algorithm}[htb]
\caption{\alg{OptimiseRules}($\globalpatterns$, $M$)}
\label{alg:pipeline}
\begin{algorithmic}
\State $\Rules_{\textnormal{candidates}} \gets \emptyset$
\For{\textbf{each} $\pattern{\varphi}$ \textbf{in} $\globalpatterns$}
    \For{\textbf{each} $\Rule$ \textbf{in} ${\mathcal{R} := \{\Rule\mid \pattern{\body{\Rule} \wedge \head{\Rule}} = \pattern{\varphi}\}}$}
        \If{$\Rule$ is term-constrained \textbf{and} $\frac{\precision{\Rule} \cdot \symmetry{\Rule}}{\prior{\Rule}} > 1$}
            \State $\Rules_{\textnormal{candidates}} \gets \Rules_{\textnormal{candidates}} \cup \{\Rule\}$
        \EndIf
    \EndFor
\EndFor
\State $\Rules_\textnormal{candidates} \gets$ \Call{chooseTopM}{$\Rules_\textnormal{candidates}$, $M$} 
\State $\Rules_\textnormal{final} \gets [ \;\, ]$ 
\While{$\Rules_{\textnormal{candidates}}$ \textbf{is not} empty} 
    \State $\Rule_{\textnormal{best}} \gets \emptyset$
    \For{\textbf{each} $\Rule$ \textbf{in} $\Rules_{\textnormal{candidates}}$}
        \If{$\utility{\{\Rule\} \cup \Rules_\textnormal{final}} > \utility{\{\Rule_{\textnormal{best}}\} \cup \Rules_\textnormal{final}}$}
            \State $\Rule_{\textnormal{best}} \gets \Rule$
        \EndIf
    \EndFor
    \State $\Rules_{\textnormal{candidates}} \gets \Rules_{\textnormal{candidates}} \setminus \{\Rule_{\textnormal{best}}\}$
    \State \textbf{append} $\Rule_{\textnormal{best}}$ to $\Rules_\textnormal{final}$
\EndWhile
\State \Return{$\Rules_\textnormal{final}$}
\end{algorithmic}
\end{algorithm}


After SPECTRUM, the rules $\Rules_{\textnormal{final}}$ can be passed to any PLM (e.g. PSL or MLN) to learn the weights of the rules (i.e. the likelihood of the rule being satisfied) for a given dataset~$\trainingdata$. We recommend adding one rule at a time to the logical model (in the order they were added to the vector $\Rules_{\textnormal{final}}$) when learning the weights, checking when the accuracy drops, as more rules may not imply a better theory. 


\subsection{Theoretical Properties}\label{section:theory}

Proofs of all results in this section can be found in App.~\ref{section:proof}. 


\paragraph{Optimality.} 

For any pattern, Alg.~\ref{alg:fragmentMining} generally finds only a subset of its ground patterns in $\datagraph$. Thus, the computed utility measures $\utilityE{\cdot}$ are estimates of the actual values $\utility{\cdot}$. To quantify the trade-off between runtime and accuracy of these estimates, we adapt the methodology proposed by Feldstein et al. \cite{prism} -- who observe that pattern occurrences in networks typically follow a Zipf distribution -- to bound the number of sampled paths, $N$, required to guarantee a maximal fractional error $\varepsilon$:

\begin{theorem}\label{theorem:N}
Suppose that pattern frequencies in $\datagraph$ are Zipfian, and let $\Rules$ be a theory whose rules' body
and rule patterns are among the $M$ most frequent patterns. To estimate the utility of $\Rules$ to within a
fractional error $\varepsilon$, it suffices to sample
\[
N = \mathcal{O}\!\left(\frac{MD\ln M}{\varepsilon^2|V|}\right)
\]
paths per node.
\end{theorem}


\begin{remark}\label{rmk:N_practice}
    In practice, we set $N = \left\lceil\frac{M\ln M}{\varepsilon^2|V|}\right\rceil$ to guarantee $\mathcal{O}(\varepsilon)$ fractional error (see \ref{rmk:N}).  
\end{remark}

\paragraph{Complexity.}
Pattern mining (Alg.~\ref{alg:fragmentMining}) follows at most $N$ paths of length at most $D$ from each
of the $|V|$ nodes, so it takes $\mathcal{O}(N|V|D)$ time in the worst case (see App.~\ref{section:pattern_complexity} for a tigher bound). Also, from Remark \ref{rmk:N_practice}, $N \leq 1 + \frac{M \ln M}{\varepsilon^2 \vert V \vert}$, hence
$N|V| \le |V| + M\ln M/\varepsilon^2$. In addition, optimising the utility costs
$\mathcal{O}(M^3(|E| + D + |\mathcal{P}|))$, where $|\mathcal{P}|$ is the number of predicates
(App.~\ref{appx:utilityComplexity}). Substituting in, the total complexity of SPECTRUM is $
\mathcal{O}\Big(|V| + \frac{MD^2\ln M}{\varepsilon^2} + M^3\big(|E| + D + |\mathcal{P}|\big)\Big)$.
Importantly, this is \emph{linear} in the number of entities $|V|$ and relations $|E|$ for a fixed $D$, $\varepsilon$, $|\mathcal{P}|$ and $M$.



\paragraph{Noise Robustness.} Thm. \ref{theorem:N} was derived assuming noiseless data. However, we can show that SPECTRUM's utility estimates, and thus the entire rule-mining pipeline, are also robust to noise:
\begin{theorem}\label{thm:noise_robustness}
Suppose each fact of the training data is independently missing with probability $p \in [0,1]$. Then, under
this noise model, for every rule set $\Rules$, the noisy utility $\tilde U(\Rules)$ is within a fractional
error $\mathcal{O}(L_{\text{max}}p)$ of the true utility $U(\Rules)$. Consequently, if SPECTRUM is run on the noisy data
with $N$ as in Thm.~\ref{theorem:N}, its utility estimates are within a fractional error
$\varepsilon + \mathcal{O}(L_{\text{max}}p)$ of the true utility.
\end{theorem}

\paragraph{Completeness.} 
%
Prior work used random walks (RWs) to mine rules \citep{lsm, cheng2023neural,prism}. Alg.~\ref{alg:fragmentMining} can be thought of as running RWs in parallel, while avoiding repeating the same walk twice. This makes Alg.~\ref{alg:fragmentMining} more efficient than naive random walks. Most importantly, RWs can miss ground patterns due to random sampling. In contrast, we can derive guarantees for the completeness of Alg.~\ref{alg:fragmentMining}:


\begin{theorem}\label{theorem:N-closeness-completness}
    Let $v$ and $v'$ be two nodes in $\datagraph$. We say that $v'$ is \notion{$N$-close} to $v$ if, for each path $(v_{i_0}, e_{i_0}, \dots, e_{i_{l-1}}, v_{i_l})$ of length $l \leq D$ between $v_{i_0} = v$ and 
        $v_{i_l} = v'$, it holds that
        $
        \vert \restrictedbinaries{v_{i_0}}{\datagraph} \vert \prod \limits_{j=1}^{l-1} (\vert \restrictedbinaries{v_{i_j}}{\datagraph} \vert - 1) \leq N.
        $ 
        Then, for each $v \in \datagraph$, SPECTRUM 
        mines all ground patterns involving $v$ and nodes that are $N$-close to $v$ and a distance $\leq D$ from $v$; 
    all other ground patterns involving $v$ and nodes within distance $D$ are found with a probability at least as large as when running $N$ RWs from $v$.
\end{theorem}

In App.~\ref{section:rw_comparison}, we elaborate on the theoretical advantages of SPECTRUM over RWs, and and show its superior performance empirically in an ablation study. While similar theoretical results regarding optimality have been derived for PRISM \citep{prism}, to the best of our knowledge, we are the first to present guarantees regarding robustness to noise as well as a complexity analysis for the entire structure-learning pipeline. 
\section{Experiments}
\label{section:experiments}

To test the performance of SPECTRUM, we compare against state-of-the-art structure learners using fuzzy logic semantics on knowledge graph completion (KGC). To demonstrate scalability, we then run SPECTRUM on two larger datasets used in PSL literature. We present ablation studies in App.~\ref{appx:ablations}.

\paragraph{Tasks.} KGC is a task commonly used by structure learning frameworks to assess the quality of learnt rules \citep{NeuralLP, DBLP:conf/nips/SadeghianADW19, rnnlogic, cheng2023neural}. 
The goal is 
to infer missing entities (i.e. given $\textsc{p}($alice$,X)$ predict $X$). For this task, we consider the five datasets: Family \citep{hinton_learning_1986}, UMLS \citep{kok_statistical_2007}, Kinship \citep{kok_statistical_2007}, WN18RR \citep{dettmers_convolutional_2018} and FB15K-237 \citep{toutanova_observed_2015}. 
To test scalability, we learn PSL rules on CAD
\citep{collective-psl} ($2.4 \times 10^5$ facts) and Yelp \citep{psl} ($2.2 \times 10^6$
facts), and measure how many of the rules hand-crafted by domain experts SPECTRUM recovers
\citep{collective-psl, psl}.
For Yelp, hand-engineered rules are provided by \cite{kouki2015hyper}, and for CAD by \cite{collective-psl}. Dataset details are in App.~\ref{app:experimentDetails}.

\paragraph{Method.} To evaluate the MRR and Hit@10, 
we used the NCRL script~\citep{cheng2023neural}. For SPECTRUM we used precision corrected for symmetry as the rule weight. Predicted entities are ranked by the sum of the weights of every rule that is satisfied with that entity in its grounding. We compare against {AMIE3~\citep{lajus_fast_2020}}, CTP \cite{minervini2020learning}, RNNlogic \citep{rnnlogic} and NCRL \citep{cheng2023neural}; further baselines are discussed in App.~\ref{sec:appx:KGC}. 
The baseline methods ran on V100 GPUs (32\,GB memory, 40 CPUs). SPECTRUM ran on a 6-core 2.60GHz i7-10750H CPU. In App. \ref{appx:sensitivityEpsilonAndM}, we present ablation studies to evaluate the effect of $M$ and $\varepsilon$. We find that SPECTRUM is insensitive to $M$ and $\varepsilon$ over a wide range, with $M=20|\mathcal{P}|$ (where $|\mathcal{P}|$ is the number of distinct predicates) and $\varepsilon=0.01$ sufficient to find all the rules discovered with larger budgets while keeping runtime close to its minimum; we therefore use these values throughout our experiments.

\begin{wrapfigure}[17]{R}{0.35\textwidth}
\centering
\includegraphics[width=0.35\textwidth]{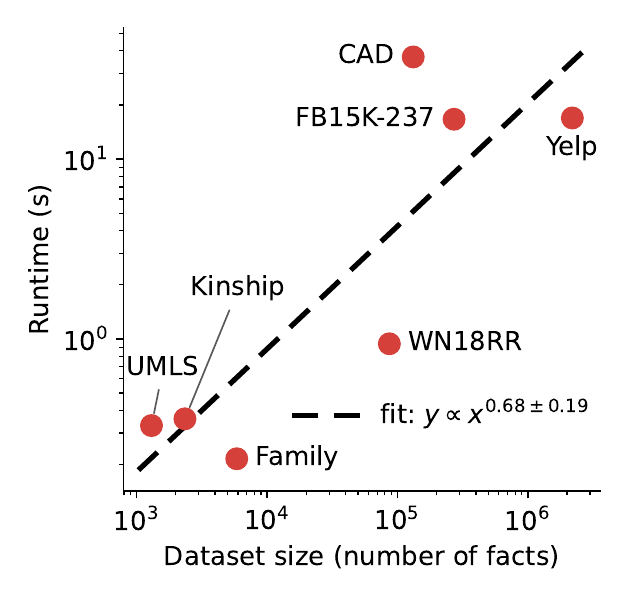}
\caption{SPECTRUM's linear scaling in dataset size.}
\label{fig:scaling}
\end{wrapfigure}

\paragraph{Results.} 
Evaluation results for KGC are shown in Tab.~\ref{tab:combined_metrics_runtimes}. With the notable exception of Kinship, \alg{SPECTRUM} outperforms neural methods in MRR and Hit@10, while consistently offering a significantly faster runtime ($100 - 1000\times$ improvement) and more efficient memory usage. Among the neural methods, NCRL usually performs best. For CAD and Yelp, \alg{SPECTRUM}'s runtimes were $37$s and $17$s respectively, and recovered (almost) exactly the logical theory created by domain experts (21/21 rules recovered for CAD, 24/26 rules recovered for YELP, see App.~\ref{appx:PSL} for details of the learnt rules). 
As shown in Fig.~\ref{fig:scaling}, we see that runtime grows at most linearly with dataset size. Further, when the size of a single dataset
is varied with all hyperparameters fixed, it is exactly linear  (App.~\ref{appx:scaling}).
Importantly, SPECTRUM can process $\mathcal{O}(10^6)$ facts notably faster than SOTA can process order $\mathcal{O}(10^3)$ facts ($\geq480$s for all baselines on Kinship vs $17$s for SPECTRUM on Yelp), 
demonstrating how SPECTRUM successfully overcomes the scalability issues of prior art.

\begin{table*}[ht!]
\caption{Performance (MRR, Hit@10, Rules recovered) and runtimes (s) for AMIE3, RNNLogic, CTP, NCRL, and SPECTRUM across datasets. Dataset sizes shown in parentheses. `$-$' indicates that an experiment did not terminate within 24 hours running on a 32\,GB V100 GPU. `$/$' indicates that the experiment ran out of RAM and crashed before completing.}
\label{tab:combined_metrics_runtimes}
\centering
{
\begin{tabular}{llccccc}
\toprule
Dataset (Size) & Metric & AMIE3 & RNNLogic & CTP & NCRL & SPECTRUM \\
\midrule

\multirow{3}{*}{\makecell[l]{Family\\($6 \times 10^3$)}} 
    & Runtime (s) & 19.5 & 1200 & 3300 & 88 & \textbf{0.2} \\
    & MRR         & 0.430 & 0.278 & 0.269 & 0.873 & \textbf{0.901} \\
    & Hit@10      & 0.766 & 0.494 & 0.542 & 0.993 & \textbf{0.999} \\

\midrule

\multirow{3}{*}{\makecell[l]{UMLS\\($1 \times 10^3$)}} 
    & Runtime (s) & 810 & 1200 & 610 & 420 & \textbf{0.3} \\
    & MRR         & 0.064 & 0.689 & 0.299 & 0.659 & \textbf{0.761} \\
    & Hit@10      & 0.161 & 0.824 & 0.703 & 0.853 & \textbf{0.952} \\

\midrule

\multirow{3}{*}{\makecell[l]{Kinship\\($2 \times 10^3$)}} 
    & Runtime (s) & 3440 & 1300 & 790 & 480 & \textbf{0.4} \\
    & MRR         & 0.168 & 0.535 & 0.306 &\textbf{0.592} & 0.451 \\
    & Hit@10      & 0.454 & \textbf{0.919} & 0.733 & 0.897 & 0.834 \\

\midrule

\multirow{3}{*}{\makecell[l]{WN18RR\\($1 \times 10^5$)}} 
    & Runtime (s) & 128 & – & -- & 2700  & \textbf{0.9} \\
    & MRR         & 0.434 & – & -- & \textbf{0.506}   & 0.500 \\
    & Hit@10      & 0.491 & – & -- & 0.687 & \textbf{0.890} \\

\midrule

\multirow{3}{*}{\makecell[l]{FB15K-237\\($3 \times 10^5$)}} 
    & Runtime (s) & 9950 & -- & -- & {/} & \textbf{16.7} \\
    & MRR         & 0.243 & -- & -- & {/} & \textbf{0.501} \\
    & Hit@10      & 0.377 & -- & -- & {/} & \textbf{0.682} \\

\bottomrule




\end{tabular}}
\end{table*}


\paragraph{Discussion.} The results for SPECTRUM in Tab.~\ref{tab:combined_metrics_runtimes} were obtained using only CPUs, whereas all baselines required a 32 GB GPU (initial attempts on a 6 GB GPU failed). 
Given that the hardware used for the baselines consumes $\sim 7\times$ more energy \citep{CPUConsumption,GPUConsumption}, SPECTRUM reduces energy consumption by more than $99.8\%$. This is increasingly important at a time when the AI industry consumes as much energy as entire countries \citep{AIConsumption}.
Furthermore, the runtime for NCRL excludes the cost of tuning its six hyperparameters to each dataset, while SPECTRUM used fixed hyperparameters on all datasets and did not require tuning. Additionally, note that NCRL ran out of memory on the FB15K-237 dataset ($\sim 10^5$ facts), implying that it would not scale to the CAD ($\sim 10^5$) or Yelp ($\sim 10^6$) datasets. We hypothesise that SPECTRUM outperforms neural models like NCRL because the latter generate a very large number of rules ($\mathcal{O}(10^3|\mathcal{P}|)$, where $|\mathcal{P}|$ is the number of unique predicates in the data), many of which are low-quality and only add noise \citep{cheng2023neural,rnnlogic}. In contrast, SPECTRUM filters for rules $\Rule$ that perform strictly better than random chance (i.e., $\precision{\Rule} \cdot \symmetry{\Rule}/\prior{\Rule} > 1$). It then ranks these by  utility, selecting at most the top $M$ to form a theory that is both compact and expressive. 
Note that we set $M$ much larger than the size of the final theory mined by \alg{SPECTRUM} (e.g. for CAD we allowed a theory of size $M=380$ but the final theory consisted of 21 rules), i.e. \alg{SPECTRUM} decided that every other pattern it found did not lead to informative rules, which is consistent with the domain expertise.

\newpage

\bibliography{references}

@article{spyropoulou2014interesting,
  title={Interesting pattern mining in multi-relational data},
  author={Spyropoulou, Eirini and De Bie, Tijl and Boley, Mario},
  journal={Data Mining and Knowledge Discovery},
  volume={28},
  pages={808--849},
  year={2014},
  publisher={Springer}
}

@inproceedings{alam2021mining,
  title={Mining frequent patterns from hypergraph databases},
  author={Alam, Md Tanvir and Ahmed, Chowdhury Farhan and Samiullah, Md and Leung, Carson K},
  booktitle={Pacific-Asia Conference on Knowledge Discovery and Data Mining},
  pages={3--15},
  year={2021},
  organization={Springer}
}

@book{cochran1977sampling,
  author    = {Cochran, William G.},
  title     = {Sampling Techniques},
  edition   = {3},
  publisher = {John Wiley \& Sons},
  address   = {New York},
  year      = {1977},
  isbn      = {978-0-471-16240-7}
}

@article{geng2006interestingness,
  title={Interestingness measures for data mining: A survey},
  author={Geng, Liqiang and Hamilton, Howard J},
  journal={ACM Computing Surveys (CSUR)},
  volume={38},
  number={3},
  pages={9--es},
  year={2006},
  publisher={ACM New York, NY, USA}
}

@article{campero_logical_2018,
  title={Logical rule induction and theory learning using neural theorem proving},
  author={Campero, Andres and Pareja, Aldo and Klinger, Tim and Tenenbaum, Josh and Riedel, Sebastian},
  journal={arXiv preprint arXiv:1809.02193},
  year={2018}
}

@InProceedings{pmlr-v162-glanois22a,
  title={Neuro-symbolic hierarchical rule induction},
  author={Glanois, Claire and Jiang, Zhaohui and Feng, Xuening and Weng, Paul and Zimmer, Matthieu and Li, Dong and Liu, Wulong and Hao, Jianye},
  booktitle={Proceedings of the 39th International Conference on Machine Learning (ICML)},
  pages={7583--7615},
  year={2022},
  organization={PMLR}
}

@inproceedings{cheng2022rlogic,
  author    = {Cheng, Kewei and Liu, Jiahao and Wang, Wei and Sun, Yizhou},
  title     = {Rlogic: Recursive Logical Rule Learning from Knowledge Graphs},
  booktitle = {Proceedings of the 28th ACM SIGKDD Conference on Knowledge Discovery and Data Mining},
  pages     = {179--189},
  year      = {2022}
}

@inproceedings{lu2022r5,
  author    = {Lu, S. and Liu, B. and Mills, K. G. and Jui, S. and Niu, D.},
  title     = {{R5}: Rule Discovery with Reinforced and Recurrent Relational Reasoning},
  booktitle = {International Conference on Learning Representations (ICLR)},
  year      = {2022}
}

@article{abstractNeSYSurvey,
  title={Neural-symbolic learning and reasoning: A survey and interpretation},
  author={d’Avila Garcez, Artur S and Bader, Sebastian and Bowman, Howard and Lamb, Luis C and de Penning, Leo and Illuminoo, BV and Poon, Hoifung},
  journal={Neuro-Symbolic Artificial Intelligence: The State of the Art},
  volume={342},
  number={1},
  pages={327},
  year={2022},
  publisher={IOS Press}
}

@inproceedings{hinton_learning_1986,
  title={Learning distributed representations of concepts},
  author={Hinton, Geoffrey E},
  booktitle={Proceedings of the Annual Meeting of the Cognitive Science Society},
  volume={8},
  year={1986}
}

@inproceedings{kok_statistical_2007,
  title={Statistical predicate invention},
  author={Kok, Stanley and Domingos, Pedro},
  booktitle={Proceedings of the 24th International Conference on Machine Learning (ICML)},
organization={PMLR},
  pages={433--440},
  year={2007}
}

@article{law2020ilasp,
  title   = {The ILASP system for learning answer set programs},
  author  = {Law, Mark and Russo, Alessandra and Broda, Krysia},
  journal = {arXiv preprint arXiv:2005.00904},
  year    = {2020}
}

@article{cropper2021learning,
  title     = {Learning logic programs by learning from failures},
  author    = {Cropper, Andrew and Duman{\v{c}}i{\'{c}}, Sebastijan},
  journal   = {Machine Learning},
  volume    = {110},
  number    = {4},
  pages     = {801--856},
  year      = {2021},
  publisher = {Springer}
}

@article{cropper2016metagol,
  title   = {Learning Higher-Order Logic Programs through Abstraction and Invention},
  author  = {Cropper, Andrew and Muggleton, Stephen H.},
  journal = {Proceedings of the Twenty-Fifth International Joint Conference on Artificial Intelligence (IJCAI)},
  pages   = {1418--1424},
  year    = {2016}
}

@article{muggleton1991inductive,
  title     = {Inductive logic programming},
  author    = {Muggleton, Stephen},
  journal   = {New Generation Computing},
  volume    = {8},
  number    = {4},
  pages     = {295--318},
  year      = {1991},
  publisher = {Springer}
}

@inproceedings{dettmers_convolutional_2018,
  title={Convolutional 2{D} knowledge graph embeddings},
  author={Dettmers, Tim and Minervini, Pasquale and Stenetorp, Pontus and Riedel, Sebastian},
  booktitle={Proceedings of the {AAAI} Conference on Artificial Intelligence},
  year={2018}
}

@inproceedings{toutanova_observed_2015,
  title={Observed versus latent features for knowledge base and text inference},
  author={Toutanova, Kristina and Chen, Danqi},
  booktitle={Proceedings of the 3rd Workshop on Continuous Vector Space Models and Their Compositionality},
  pages={57--66},
  year={2015}
}

@inproceedings{mao2019neuro,
  title={The neuro-symbolic concept learner: Interpreting scenes, words, and sentences from natural supervision},
  author={Mao, Jiayuan and Gan, Chuang and Kohli, Pushmeet and Tenenbaum, Joshua B and Wu, Jiajun},
  booktitle={The 7th International Conference on Learning Representations (ICLR)},
  year={2019},
  publisher={OpenReview.net}
}

@inproceedings{gu2019scene,
  title={Scene graph generation with external knowledge and image reconstruction},
  author={Gu, Jiuxiang and Zhao, Handong and Lin, Zhe and Li, Sheng and Cai, Jianfei and Ling, Mingyang},
  booktitle={Proceedings of the IEEE/CVF Conference on Computer Vision and Pattern Recognition (CVPR)},
  pages={1969--1978},
  publisher    = {{IEEE} Computer Society},
  year={2019}
}

@article{bomze_maximum_1999,
  title={The maximum clique problem},
  author={Bomze, Immanuel M and Budinich, Marco and Pardalos, Panos M and Pelillo, Marcello},
  journal={Handbook of Combinatorial Optimization: Supplement Volume A},
  pages={1--74},
  year={1999},
  publisher={Springer}
}

@inproceedings{lsm,
  title={Learning {M}arkov logic networks using structural motifs},
  author={Kok, Stanley and Domingos, Pedro},
  booktitle={Proceedings of the 27th International Conference on Machine Learning (ICML)},
  pages={551--558},
organization={PMLR},
  year={2010}
}

@article{psl,
  author    = {Bach, Stephen H and Broecheler, Matthias and Huang, Bert and Getoor, Lise},
  title     = {Hinge-Loss {M}arkov Random Fields and Probabilistic Soft Logic},
  journal   = {Journal of Machine Learning Research},
  volume    = {18},
  year      = {2017}
}

@article{MLN,
  title={Markov logic networks},
  author={Richardson, Matthew and Domingos, Pedro},
  journal={Machine Learning},
  volume={62},
  number={1},
  pages={107--136},
  year={2006},
  publisher={Springer}
}

@inproceedings{mihalkova_bottom-up_2007,
  title={Bottom-up learning of Markov logic network structure},
  author={Mihalkova, Lilyana and Mooney, Raymond J},
  booktitle={Proceedings of the 24th International Conference on Machine Learning (ICML)},
  pages={625--632},
organization={PMLR},
  year={2007}
}

@article{deepproblog,
  title={Deepproblog: Neural probabilistic logic programming},
  author={Manhaeve, Robin and Dumancic, Sebastijan and Kimmig, Angelika and Demeester, Thomas and De Raedt, Luc},
  journal={Advances in Neural Information Processing Systems (NeurIPS)},
  volume={31},
  year={2018}
}

@inproceedings{lajus_fast_2020,
  title={Fast and exact rule mining with {AMIE} 3},
  author={Lajus, Jonathan and Gal{\'a}rraga, Luis and Suchanek, Fabian},
  booktitle={The Semantic Web: 17th International Conference, ESWC 2020, Heraklion, Crete, Greece, May 31--June 4, 2020, Proceedings 17},
  pages={36--52},
  year={2020},
  organization={Springer}
}

@article{gao_differentiable_2024,
  title={A differentiable first-order rule learner for inductive logic programming},
  author={Gao, Kun and Inoue, Katsumi and Cao, Yongzhi and Wang, Hanpin},
  journal={Artificial Intelligence},
  volume={331},
  pages={104108},
  year={2024},
  publisher={Elsevier}
}

@article{FOIL,
  title={Learning logical definitions from relations},
  author={Quinlan, J. Ross},
  journal={Machine Learning},
  volume={5},
  pages={239--266},
  year={1990},
  publisher={Springer}
}

@article{mdie,
  title={Inverse entailment and {P}rogol},
  author={Muggleton, Stephen H},
  journal={New Generation Computing},
  volume={13},
  pages={245--286},
  year={1995},
  publisher={Springer}
}

@article{Inspire,
  title={Best-effort inductive logic programming via fine-grained cost-based hypothesis generation: The inspire system at the inductive logic programming competition},
  author={Sch{\"u}ller, Peter and Benz, Mishal},
  journal={Machine Learning},
  volume={107},
  pages={1141--1169},
  year={2018},
  publisher={Springer}
}

@article{diffILP,
  author    = {Richard Evans and
               Edward Grefenstette},
  title     = {Learning Explanatory Rules from Noisy Data},
  journal   = {Journal of Artificial Intelligence Research},
  volume    = {61},
  pages     = {1--64},
  year      = {2018}
}

@article{DBLP:conf/nips/SadeghianADW19,
  title={{DRUM}: End-to-end differentiable rule mining on knowledge graphs},
  author={Sadeghian, Ali and Armandpour, Mohammadreza and Ding, Patrick and Wang, Daisy Zhe},
  journal={Advances in Neural Information Processing Systems (NeurIPS)},
  volume={32},
  year={2019}
}

@online{cropper2021inductive,
  author       = {Cropper, Andrew and Duman{\v{c}}i{\'{c}}, Sebastijan and Evans, Richard and Muggleton, Stephen H.},
  title        = {Inductive logic programming at 30},
  year         = {2021},
  eprint       = {2102.10556},
  archiveprefix = {arXiv},
  primaryclass = {cs.AI},
  doi          = {10.48550/arXiv.2102.10556},
  pubstate     = {prepublished}
}

@article{boostr,
  title={Gradient-based boosting for statistical relational learning: the Markov logic network and missing data cases},
  author={Khot, Tushar and Natarajan, Sriraam and Kersting, Kristian and Shavlik, Jude},
  journal={Machine Learning},
  volume={100},
  number={1},
  pages={75--100},
  year={2015},
  publisher={Springer}
}

@article{friedman2000greedy,
  author = {Friedman, Jerome H.},
  journal = {Annals of Statistics},
  pages = {1189--1232},
  title = {Greedy Function Approximation: A Gradient Boosting Machine},
  volume = 29,
  year = 2000
}

@inproceedings{Sen_Carvalho_Riegel_Gray_2022, 
    title={Neuro-Symbolic Inductive Logic Programming with Logical Neural Networks}, 
  booktitle={Proceedings of the 36th AAAI Conference on Artificial Intelligence},
  organization={Association for the Advancement of Artificial Intelligence},
    author={Sen, Prithviraj and Carvalho, Breno W. S. R. de and Riegel, Ryan and Gray, Alexander}, 
    year={2022}, 
    pages={8212-8219} 
}

@inproceedings{meta-abd,
  title     = {Abductive Knowledge Induction from Raw Data},
  author    = {Dai, Wang-Zhou and Muggleton, Stephen},
  booktitle={Proceedings of the 30th International Joint Conference on Artificial Intelligence (IJCAI)},
  year={2021},
  organization={{IJCAI}}
}

@inproceedings{NLMs,
  author    = {Honghua Dong and
               Jiayuan Mao and
               Tian Lin and
               Chong Wang and
               Lihong Li and
               Denny Zhou},
  title     = {Neural Logic Machines},
  booktitle = {The 7th International Conference on Learning Representations (ICLR)},
  publisher = {OpenReview.net},
  year      = {2019}
}

@article{NeuralLP,
  title={Differentiable learning of logical rules for knowledge base reasoning},
  author={Yang, Fan and Yang, Zhilin and Cohen, William W},
  journal={Advances in Neural Information Processing Systems (NeurIPS)},
  volume={30},
  year={2017}
}

@article{LNNs,
  author       = {Ryan Riegel and
                  Alexander G. Gray and
                  Francois P. S. Luus and
                  Naweed Khan and
                  Ndivhuwo Makondo and
                  Ismail Yunus Akhalwaya and
                  Haifeng Qian and
                  Ronald Fagin and
                  Francisco Barahona and
                  Udit Sharma and
                  Shajith Ikbal and
                  Hima Karanam and
                  Sumit Neelam and
                  Ankita Likhyani and
                  Santosh K. Srivastava},
  title        = {Logical Neural Networks},
  journal      = {CoRR},
  volume       = {abs/2006.13155},
  year         = {2020}
}

@inproceedings{prism,
  title={Principled and efficient motif finding for structure learning of lifted graphical models},
  author={Feldstein, Jonathan and Phillips, Dominic and Tsamoura, Efthymia},
  booktitle={Proceedings of the 37th AAAI Conference on Artificial Intelligence},
  volume={37},
  year={2023},
  organization={Association for the Advancement of Artificial Intelligence}
}

@inproceedings{cheng2023neural,
  title={Neural Compositional Rule Learning for Knowledge Graph Reasoning},
  author={Cheng, Kewei and Ahmed, Nesreen K. and Sun, Yizhou},
  booktitle={The 11th International Conference on Learning Representations (ICLR)},
  publisher    = {OpenReview.net},
  year={2023}
}

@inproceedings{rnnlogic,
  title={{RNNL}ogic: Learning Logic Rules for Reasoning on Knowledge Graphs},
  author={Qu, Meng and Chen, Junkun and Xhonneux, Louis-Pascal and Bengio, Yoshua and Tang, Jian},
  booktitle    = {The 9th International Conference on Learning Representations,(ICLR)},
  publisher    = {OpenReview.net},
  year={2021}
}

@article{zimmer2023differentiable,
  author    = {Zimmer, Matthieu and Feng, Xuening and Glanois, Claire and Jiang, Zhaohui and Zhang, Jianyi and Weng, Paul and Li, Dong and Hao, Jianye and Liu, Wulong},
  title     = {Differentiable Logic Machines},
  journal   = {Transactions on Machine Learning Research},
  year      = {2023},
  url       = {https://openreview.net/forum?id=mXfkKtu5JA},
  note      = {2023}
}

@inproceedings{zhang2019iteratively,
  author    = {Zhang, Wen and Paudel, Bibek and Wang, Liang and Chen, Jiaoyan and Zhu, Hai and Zhang, Wei and Bernstein, Abraham and Chen, Huajun},
  title     = {Iteratively Learning Embeddings and Rules for Knowledge Graph Reasoning},
  booktitle = {Proceedings of the World Wide Web Conference (WWW)},
  pages     = {2366--2377},
  year      = {2019},
  publisher = {ACM},
  doi       = {10.1145/3308558.3313612}
}

@inproceedings{concordia,
  title={Parallel neurosymbolic integration with {C}oncordia},
  author={Feldstein, Jonathan and Jur{\v{c}}ius, Modestas and Tsamoura, Efthymia},
  booktitle={Proceedings of the 40th International Conference on Machine Learning (ICML)},
  pages={9870--9885},
  year={2023},
  publisher={PMLR}
}

@inproceedings{deraedt2015inducing,
  author    = {De Raedt, Luc and Dries, Anton and Thon, Ingo and Van den Broeck, Guy and Verbeke, Mathias},
  title     = {Inducing Probabilistic Relational Rules from Probabilistic Examples},
  booktitle = {Proceedings of the 24th International Joint Conference on Artificial Intelligence (IJCAI)},
  pages     = {1835--1842},
  year      = {2015},
  publisher = {AAAI Press}
}

@article{shindo2023alphailp,
  author    = {Shindo, Hikaru and Pfanschilling, Viktor and Dhami, Devendra Singh and Kersting, Kristian},
  title     = {{$\alpha$ILP}: Thinking Visual Scenes as Differentiable Logic Programs},
  journal   = {Machine Learning},
  volume    = {112},
  number    = {5},
  year      = {2023},
  pages     = {1465--1494},
  publisher = {Springer},
  doi       = {10.1007/s10994-023-06311-6}
}

@inproceedings{minervini2020learning,
  author    = {Minervini, Pasquale and Riedel, Sebastian and Stenetorp, Pontus and Grefenstette, Edward and Rockt{\"a}schel, Tim},
  title     = {Learning Reasoning Strategies in End-to-End Differentiable Proving},
  booktitle = {Proceedings of the 37th International Conference on Machine Learning (ICML)},
  pages     = {6938--6949},
  year      = {2020},
  publisher = {PMLR}
}

@inproceedings{kouki2015hyper,
  author       = {Pigi Kouki and
                  Shobeir Fakhraei and
                  James R. Foulds and
                  Magdalini Eirinaki and
                  Lise Getoor},
  title        = {Hy{PER}: {A} Flexible and Extensible Probabilistic Framework for Hybrid
                  Recommender Systems},
  booktitle    = {Proceedings of the 9th {ACM} Conference on Recommender Systems (RecSys)},
  pages        = {99--106},
  publisher    = {{ACM}},
  year         = {2015}
}

@Incollection{Kakas2017,
author="Kakas, Antonis C.",
title="Abduction",
booktitle="Encyclopedia of Machine Learning and Data Mining",
year="2017",
publisher="Springer US",
address="Boston, MA",
pages="1--8"
}

@INPROCEEDINGS{collective-psl,
  title={Collective activity detection using hinge-loss Markov random fields},
  author={London, Ben and Khamis, Sameh and Bach, Stephen and Huang, Bert and Getoor, Lise and Davis, Larry},
  booktitle={Proceedings of the IEEE Conference on Computer Vision and Pattern Recognition (CVPR) Workshops},
  publisher    = {{IEEE} Computer Society},
  pages={566--571},
  year={2013}
}

@article{feldstein2024mapping,
  title={Mapping the neuro-symbolic {AI} landscape by architectures: A handbook on augmenting deep learning through symbolic reasoning},
  author={Feldstein, Jonathan and Dilkas, Paulius and Belle, Vaishak and Tsamoura, Efthymia},
  journal={arXiv preprint arXiv:2410.22077},
  year={2024}
}

@misc{GPUConsumption,
author={Nvidia},
  title = {NVIDIA V100
TENSOR CORE GPU},
  howpublished = {\url{https://images.nvidia.com/content/technologies/volta/pdf/volta-v100-datasheet-update-us-1165301-r5.pdf}},
  year = {2020}
}

@misc{CPUConsumption,
title = {Intel Core i7-10750H},
author={Intel},
  howpublished = {\url{https://www.intel.com/content/www/us/en/products/sku/201837/intel-core-i710750h-processor-12m-cache-up-to-5-00-ghz/specifications.html}},
  year = {2020}
}

@misc{AIConsumption,
title = {Warning {AI} industry could use as much energy as the Netherlands},
author={Kleinman, Zoe and Vallance, Chris},
  howpublished = {\url{  https://www.bbc.co.uk/news/technology-67053139}},
  year = {2023}
}
\bibliographystyle{plain}

\newpage
\appendix

\section{Notation}\label{app:notation}

Table~\ref{tab:notation} lists the notation of the main text, grouped by where it
is introduced.

\begin{table}[!ht]
\caption{Notation.}
\label{tab:notation}
\centering
\small
\renewcommand{\arraystretch}{1.0}
\begin{tabular}{@{}l p{0.62\linewidth} l@{}}
\toprule
Symbol & Meaning & Introduced \\
\midrule
\multicolumn{3}{@{}l}{\emph{Logic and data}} \\
$\trainingdata$ & Relational database: a set of facts & Sec.~\ref{section:prelims} \\
$\constant{c}$;\; $X, Y, Z$ & Constant (an object of the domain); variable & Sec.~\ref{section:prelims} \\
$\pred(t_1, \dots, t_n)$ & Atom: an $n$-ary predicate $\pred$ applied to terms $t_i$ (constants or variables) & Sec.~\ref{section:prelims} \\
$\atom$;\; $\logicfact$ & Atom; fact, i.e.\ a ground atom & Sec.~\ref{section:prelims} \\
$\groundatoms$ & Set of groundings of $\atom$ in $\trainingdata$ & Sec.~\ref{section:prelims} \\
$\mathcal{P}$ & Set of predicates in $\trainingdata$ & Sec.~\ref{section:utility} \\
$\logicformula$ & Conjunction of atoms & Sec.~\ref{section:utility} \\
$\Rule$;\; $\body{\Rule}$, $\head{\Rule}$ & Rule; its body and head & Sec.~\ref{section:prelims} \\
$L(\cdot)$ & Length of a conjunction or rule: its number of atoms & Sec.~\ref{section:prelims} \\
$\Rules$;\; $\Rules_\atom$ & Theory (a set of rules); the rules in $\Rules$ with head $\atom$ & Sec.~\ref{section:prelims} \\
$\atoms$ & Set of head atoms of $\Rules$ & Def.~\ref{definition:utilityRuleSet} \\
\midrule
\multicolumn{3}{@{}l}{\emph{Graphs and patterns}} \\
$\graph = (V, E)$ & Hypergraph with nodes $V$ and edges $E$, each edge a set of nodes & Sec.~\ref{section:prelims} \\
$\datagraph$ & Datagraph of $\trainingdata$: a node per constant, an edge labelled $\pred$ per fact $\pred(\constant{c}_1, \dots, \constant{c}_k)$; $|V|$ entities, $|E|$ facts & Sec.~\ref{section:prelims} \\
$v, v', v_0$;\; $e$ & Nodes; an edge & Sec.~\ref{section:prelims} \\
$\restrictedunaries{v}{\graph}$, $\restrictedbinaries{v}{\graph}$ & Unary and binary edges of $\graph$ incident to $v$ & Sec.~\ref{section:pattern mining} \\
$\pattern{\logicformula}$ & Pattern of $\logicformula$: a node per term, an edge per atom. $\pattern{\body{\Rule} \wedge \head{\Rule}}$ is the rule pattern of $\Rule$ & Sec.~\ref{section:utility} \\
$\underline{\mathcal{G}}$ & Ground pattern: a grounding of a pattern that is satisfied in $\trainingdata$ & Alg.~\ref{alg:fragmentMining} \\
$\instances{\logicformula}$ & Set of ground patterns of $\logicformula$ in $\trainingdata$ & Sec.~\ref{section:utility} \\
$\restrictedinstances{\logicfact}$ & Ground patterns of the rule pattern of $\Rule$ whose head is grounded to $\logicfact$ & Sec.~\ref{section:utility} \\
$\graph \circ e$ & Graph $\graph$ with edge $e$ and its nodes added & Sec.~\ref{section:pattern mining} \\
\midrule
\multicolumn{3}{@{}l}{\emph{Utility}} \\
$\precision{\Rule}$ & Precision & Def.~\ref{definition:precision} \\
$\symmetry{\Rule}$ & Symmetry factor & Def.~\ref{definition:symmetry} \\
$\prior{\Rule}$;\; $\mathcal{A}$ & Bayesian prior; atoms over all predicates with the same terms as $\head{\Rule}$ & Def.~\ref{definition:prior} \\
$\recall{\Rule}$, $\recall{\Rules_\atom}$ & Recall of a rule; of a set of rules with head $\atom$ & Defs.~\ref{definition:log-recall-rule}, \ref{definition:rule-set-recall} \\
$\complexity{\Rule}$, $\complexity{\Rules_\atom}$ & Complexity factor $e^{-L(\Rule)}$; its geometric mean over $\Rules_\atom$ & Defs.~\ref{definition:complexity}, \ref{definition:rule-set-recall} \\
$\utility{\Rule}$, $\utility{\Rules}$ & Utility of a rule; of a theory & Defs.~\ref{definition:utilityRule}, \ref{definition:utilityRuleSet} \\
$\precisionest{\Rule}$, $\utilityE{\Rules}$ & Estimates of precision and utility from the mined patterns & Sec.~\ref{section:theory} \\
$\tilde U(\Rules)$ & Utility estimate on noisy data & Thm.~\ref{thm:noise_robustness} \\
\midrule
\multicolumn{3}{@{}l}{\emph{SPECTRUM}} \\
$M$ & Maximum number of rules in the learnt theory & Sec.~\ref{sec:SPECTRUM} \\
$D$ & Mining depth: maximum number of binary edges on a mined path & Sec.~\ref{sec:SPECTRUM} \\
$L_{max} = 2D + 1$ & Maximum length of a mined pattern, and hence of a rule & Sec.~\ref{sec:SPECTRUM} \\
$\varepsilon$ & Target fractional uncertainty of the utility estimates & Sec.~\ref{sec:SPECTRUM} \\
$N$ & Path budget per start node, set from $M$ and $\varepsilon$ & Sec.~\ref{section:pattern mining} \\
$n$, $n'$;\; $d$ & Budget at the current node, and passed to each next node; current depth & Alg.~\ref{alg:fragmentMining} \\
$\edges'$;\; $\previousedges$ & Binary edges selected at the current node; edges already used on the path & Alg.~\ref{alg:fragmentMining} \\
$\currentpatterns$, $\newpatterns$, $\finalpatterns$ & Ground patterns received by a call; after unary grafting at $v$; after binary grafting, passed to the next call & Alg.~\ref{alg:fragmentMining} \\
$\globalpatterns$ & All mined ground patterns, indexed by pattern & Alg.~\ref{alg:fragmentMining} \\
$\mathcal{R}$ & Rules whose rule pattern is a given mined pattern $\pattern{\logicformula}$ & Alg.~\ref{alg:pipeline} \\
$\Rules_{\textnormal{candidates}}$;\; $\Rules_{\textnormal{final}}$;\; $\Rule_{\textnormal{best}}$ & Candidate rules; output vector of rules; rule added in the current iteration & Alg.~\ref{alg:pipeline} \\
\bottomrule
\end{tabular}
\end{table}

\section{Utility Metric Details}\label{app:design_choices}


Measuring the ``usefulness'' of rules in propositional logic has a rich history, with various probability metrics proposed \cite{geng2006interestingness}. Here, we focus on designing utility metrics for First-Order Logic (FOL). Further, we explicitly define the utility in terms of mined pattern counts (rather than probabilities) because they are the quantities directly available from the pattern mining algorithm. We motivate each of our design choices below.

\paragraph{Precision.} 
Precision serves as a computationally feasible proxy for accuracy, defined as the ratio of times the head is true given the body is true. Calculating exact rule accuracy is intractable, as it requires enumerating every ``absence'' (instances where the body is false). Our definition (once corrected for symmetry, see below) is equivalent to the definition of \textit{precision} in \cite{gao_differentiable_2024}, \textit{confidence} in \cite{lajus_fast_2020}, and \textit{confidence/precision} in \cite{geng2006interestingness}. However, unlike \cite{geng2006interestingness}, which uses a probabilistic definition ($P(\head{\Rule}|\body{\Rule})$), we formulate precision using empirical counts to match our mining outputs.

\paragraph{Symmetry factor.} This factor is a necessary correction for the counting formalism used in our precision metric. While probability-based definitions naturally account for symmetry \cite{geng2006interestingness}, count-based definitions do not. This correction is essential to avoid bias and was notably missing in \cite{gao_differentiable_2024} and \cite{lajus_fast_2020}. 

\paragraph{Bayesian prior.} This prior corrects for bias in unbalanced datasets. Without it, a rule might appear accurate simply by random chance, depending on the frequency of specific facts in the training data. The ratio in Definition \ref{definition:prior} corrects for this statistical effect, allowing us to evaluate a rule's predictive power relative to a random baseline.

\paragraph{Recall.} Recall measures how often a rule appears in the data, similar to the concept of support ($P(\body{\Rule}, \head{\Rule})$) in \cite{geng2006interestingness}. However, we introduce a specific weighting scheme: counts of rules predicting the same fact are weighted logarithmically, while those predicting different facts are weighted linearly. This ensures that rules explaining diverse data are valued higher than those repeatedly predicting facts about a single high-degree entity (e.g., an ``influencer'' in a social network). This distinction makes our metric more robust for non-homogeneous data than naive support. 


\paragraph{Complexity factor.} Recall increases exponentially with rule length. This can be seen by considering iteratively constructing a ground pattern by making decisions at each vertex to select from the set of possible valid edges to add next. Computing the total number of possibilities, it can be shown that the number of possible groundings over the whole dataset, and thus the recall associated with a rule, grows exponentially with rule length. We apply an exponential complexity factor to correct for this. Without it, utility would be biased towards preferring longer rules simply because they (on average) appear more frequently in the data.

\section{Proofs of the Theoretical Properties of SPECTRUM}
\label{section:proof}

\begin{table}[!ht]
\caption{Additional notation used in the proofs.}
\label{tab:proofNotation}
\centering
\small
\renewcommand{\arraystretch}{1.0}
\begin{tabular}{@{}l p{0.62\linewidth} l@{}}
\toprule
Symbol & Meaning & Introduced \\
\midrule
$\theta$ & A single ground pattern & App.~\ref{section:proof} \\
$c(\logicformula)$, $\hat c(\logicformula)$ & Number of ground patterns of $\logicformula$, $|\instances{\logicformula}|$; its estimate & App.~\ref{section:proof} \\
$T = |V| N$ & Total number of sampled paths & App.~\ref{section:proof:sampleComplexity} \\
$h(\theta)$;\; $\lambda(\theta)$ & Number of paths that traverse $\theta$; probability that one path from a uniformly random node traverses $\theta$ & App.~\ref{section:proof:sampleComplexity} \\
$\lambda_{\min}$, $\lambda_{\max}$;\; $d_{\max}$ & Smallest and largest $\lambda(\theta)$; maximum binary degree & App.~\ref{section:proof:sampleComplexity} \\
$q(\logicformula)$ & Probability that a path is a ground pattern of $\logicformula$, $\sum_{\theta \in \instances{\logicformula}} \lambda(\theta)$ & App.~\ref{section:proof:sampleComplexity} \\
$\mathbf{G}_l$;\; $H_l$, $H$ & Distinct patterns of length $l$; harmonic number $\sum_{m=1}^{|\mathbf{G}_l|} 1/m$; $\max_{l \le L_{\text{max}}} H_l$ & App.~\ref{section:proof:sampleComplexity} \\
$d(\logicfact)$;\; $w(\theta)$ & Number of ground patterns of $\Rules_\atom$ entailing $\logicfact$; $\ln(1 + d)/d$ & App.~\ref{section:proof:sampleComplexity} \\
$\mathcal{N}_i(v)$ & Nodes reached after $i$ steps from $v$ & Thm.~\ref{thm:complexity} \\
$p$ & Probability that a fact is deleted (noise model) & App.~\ref{section:proof:noise} \\
\bottomrule
\end{tabular}
\end{table}


\paragraph{$\varepsilon$-uncertainty.} An estimator $\hat X$ of a quantity $X > 0$ is
\emph{$\varepsilon$-uncertain} if its relative root-mean-square error is at most $\varepsilon$:
$\sqrt{\mathbb{E}[(\hat X - X)^2]} \,/\, X \le \varepsilon$. By Chebyshev's inequality, this implies
$|\hat X - X|/X < \varepsilon/\sqrt{\delta}$ with probability at least $1 - \delta$.

\subsection{Completeness}
\label{section:proofs:N-closeness-completness}

\begin{manualtheorem}{\ref{theorem:N-closeness-completness}}[Completeness]
Let $v$ and $v'$ be two nodes of $\datagraph$ a distance $l \ge 0$ apart. We say that $v'$ is
\notion{$N$-close} to $v$ if every length-$l$ path $(v_{i_0}, e_{i_0}, \dots, e_{i_{l-1}}, v_{i_l})$
from $v_{i_0} = v$ to $v_{i_l} = v'$ satisfies
\begin{equation}\label{eqn:Nclose}
|\restrictedbinaries{v_{i_0}}{\datagraph}| \prod_{j=1}^{l-1} \big(|\restrictedbinaries{v_{i_j}}{\datagraph}| - 1\big) \le N.
\end{equation}
Then, for each $v \in \datagraph$, Alg.~\ref{alg:fragmentMining} mines \emph{all} ground patterns
involving $v$ and nodes within distance $D$ that are $N$-close to $v$. Every other ground pattern
involving $v$ and nodes within distance $D$ is mined with probability at least as large as when running
$N$ random walks from $v$.
\end{manualtheorem}

\noindent\emph{Intuition.} Algorithm~\ref{alg:fragmentMining} divides its budget $N$ among the edges
it can follow. As long as every edge receives a budget of at least one, it explores all of them, so it
enumerates everything within reach. Condition~\eqref{eqn:Nclose} is exactly the condition for the budget
never to run out along any path to $v'$.

\begin{proof}
\emph{Exhaustive part.} Consider a path $(v_{i_0}, e_{i_0}, \dots, e_{i_{l-1}}, v_{i_l})$ with
$l \le D$. At the source $v_{i_0}$, the algorithm considers at most $|\restrictedbinaries{v_{i_0}}{\datagraph}|$
edges. If $N$ is at least this number, every edge, including $e_{i_0}$, is selected with a budget of at
least $\lfloor N / |\restrictedbinaries{v_{i_0}}{\datagraph}| \rfloor \ge 1$
(lines~\ref{line:binaryselection1}--\ref{line:binaryselection2}). At each later node $v_{i_j}$, the edge
the path arrived by is excluded (line~\ref{line:binaryselection1}), so at most
$|\restrictedbinaries{v_{i_j}}{\datagraph}| - 1$ edges are considered. By induction on $j$, every edge of
the path is selected whenever~\eqref{eqn:Nclose} holds. If $v'$ is $N$-close to $v$, this holds for
every path from $v$ to $v'$, so every such path is traversed. Every pattern mined by
Alg.~\ref{alg:fragmentMining} is a path or a cycle with at most one unary atom attached to each node,
and the unary atoms of every traversed node are recorded. Hence every ground pattern on these nodes is mined.

\emph{Sampled part.} Once the budget along a branch drops to one, the algorithm follows a single path,
choosing each next edge uniformly among the edges not yet used. A path of distinct edges
$(e_{i_0}, \dots, e_{i_{l-1}})$ is then followed with probability at least
$|\restrictedbinaries{v_{i_0}}{\datagraph}|^{-1} \prod_{j \ge 1} (|\restrictedbinaries{v_{i_j}}{\datagraph}| - 1)^{-1}$,
which is at least the probability $\prod_{j \ge 0} |\restrictedbinaries{v_{i_j}}{\datagraph}|^{-1}$ that
a random walk follows it, since a random walk may also step back along the edge it arrived by. Moreover,
the $N$ budget units leave the source along distinct edges (sampling without replacement), whereas $N$
independent random walks may repeat edges. Both effects can only increase the probability of covering a
given path, so any ground pattern is found with probability at least as large as with $N$ random walks.
\end{proof}

\subsection{Complexity of the Pattern Mining Algorithm}
\label{section:pattern_complexity}

\begin{theorem}[Complexity]
\label{thm:complexity}
The number of recursive calls made by Algorithm~\ref{alg:fragmentMining} is at most
\begin{equation}\label{eqn:minFunction}
\sum_{v \in \datagraph} \min\Big(|\restrictedbinaries{v}{\datagraph}| + \sum_{i=1}^{D-1} \sum_{v' \in \mathcal{N}_i(v)} \big(|\restrictedbinaries{v'}{\datagraph}| - 1\big),\; ND\Big),
\end{equation}
where $\mathcal{N}_i(v)$ is the set of nodes reached after $i$ steps from $v$. The worst-case runtime is
therefore $\mathcal{O}(|V|ND)$, and it is smaller on graphs of low binary degree.
\end{theorem}

\begin{proof}
Fix a source node $v$. Each recursive call extends the current path by one edge, so the number of calls
is the number of edge extensions the algorithm performs. Two bounds hold simultaneously.
(i) Every extension uses a distinct edge leaving a node at depth $i < D$: at most
$|\restrictedbinaries{v}{\datagraph}|$ edges at the source and at most $|\restrictedbinaries{v'}{\datagraph}| - 1$
at every later node $v'$, which is the first argument of the minimum.
(ii) The budget $N$ is split between the edges taken at each step, and every branch receives at least one
unit, so at most $N$ branches are active at each of the $D$ depths, giving at most $ND$ calls.
The number of calls is at most both bounds, hence at most their minimum. Summing over source nodes
gives~\eqref{eqn:minFunction}.
\end{proof}

\begin{remark}
Theorem~\ref{thm:complexity} counts recursive calls. With unary predicates, each call also attaches the
unary atoms of the current node to the patterns it carries, so the work per call grows with the number of
unary predicates per node. Reading the data takes a further $\mathcal{O}(|E|)$.
\end{remark}

\subsection{Optimality}
\label{section:proof:sampleComplexity}

We now show that $N = \mathcal{O}\big(M D \ln M / (\varepsilon^2 |V|)\big)$ paths per node
suffice for the utility estimates to be $\varepsilon$-uncertain. We first bound pattern count uncertainties  (Lemma~\ref{lem:counts}), then we bound the number of patterns (Lemma~\ref{lem:numberOfPatterns}), then apply these results to bound precision estimates
(Lemma~\ref{lem:precision}), recall estimates (Lemma~\ref{lem:recall}), and finally utility (Theorem~\ref{theorem:N}).

\paragraph{Preliminaries.} Suppose Algorithm~\ref{alg:fragmentMining} is run from every node, $N$
times each, $T = |V| N$ paths in total. A path \emph{traverses} a ground pattern $\theta$ of length $l$ if
its first $l$ steps, with the unary atoms attached, are exactly the atoms of $\theta$. Let $h(\theta)$ be the
number of the $T$ paths that traverse $\theta$, i.e.\ how many times $\theta$ was found, and let
$\lambda(\theta)$ be the probability that a single path, started at a uniformly random node, traverses
$\theta$. Supposing $\lambda(\theta)$ is known, we can estimate $c(\varphi)$ by
\[
\hat c(\varphi) = \sum_{\theta \in \instances{\varphi}} \frac{h(\theta)}{T \lambda(\theta)},
\]
which is unbiased, since $\mathbb{E}[h(\theta)] = T\lambda(\theta)$.
For a rule $\Rule$, the utility depends on the data only through two conjunctions:
$\varphi = \body{\Rule}$ and $\varphi = \body{\Rule} \wedge \head{\Rule}$. The precision
(Definition~\ref{definition:precision}) is the ratio of their counts,
$c(\body{\Rule} \wedge \head{\Rule}) / c(\body{\Rule})$, multiplied by the symmetry factor and the prior.
The recall (Definition~\ref{definition:log-recall-rule}) is a sum over the ground patterns
$\theta \in \instances{\body{\Rule} \wedge \head{\Rule}}$, grouped by the head fact $\logicfact$ that each one grounds.
In short, we can obtain \textit{estimates} of precision, recall and utility by substituting the count estimates into the appropriate formulae. Note that the symmetry factor, complexity factor and Bayesian prior can be computed exactly: the first two depend only
on the shape of the rule, and the prior requires one linear pass over the data (which is done anyway).

Let $d_{\max} = \max_{v} |\restrictedbinaries{v}{\datagraph}|$ be the maximum binary degree, and write $\lambda_{\min}$
and $\lambda_{\max}$ for the smallest and largest sampling probabilities among the ground patterns in the datagraph. Each of the at most $D$ binary steps of a path chooses among at most $d_{\max}$ edges, while
unary atoms are attached deterministically, so for any two ground patterns of the same length
\begin{equation}\label{eqn:samplingRatio}
\lambda_{\max} / \lambda_{\min} \le d_{\max}^{D}.
\end{equation}
This is a worst case: the ratio equals one when all ground patterns are found with certainty
(Theorem~\ref{theorem:N-closeness-completness}), and it is large only around high-degree hubs.

\paragraph{Assumption.} In what follows, we make one assumption on the statistics of the data:
\begin{enumerate}
\item[(A1)] \emph{Zipfian patterns.} For each length $l$, let $q(\varphi) = \sum_{\theta \in \instances{\varphi}} \lambda(\theta)$
be the probability that a path of length $l$, started at a uniformly random node, is a ground pattern of
$\varphi$. Following Feldstein et al. \citep{prism}, we will assume that these probabilities follow Zipf's law: the $k$-th most probable pattern has $q = 1/(k H_l)$, where
$H_l = \sum_{m=1}^{|\mathbf{G}_l|} 1/m$.
\end{enumerate}

\begin{lemma}[Count uncertainty]\label{lem:counts}
Under (A1), if $|V| N \ge M H_l / \varepsilon'^2$, then $\hat c(\varphi)$ is $d_{\max}^{D/2}\varepsilon'$-uncertain for
each of the $M$ most frequent patterns of length $l$.
\end{lemma}

\begin{proof}
First suppose the $T$ paths start at independent, uniformly random nodes. A path has a single prefix of
length $l$, so it traverses at most one ground pattern of length $l$. Each path therefore falls into
category $\theta$ with probability $\lambda(\theta)$, independently of the others, and the counts
$h(\theta)$ are multinomial with $T$ trials:
\[
\mathrm{Var}[h(\theta)] = T\lambda(\theta)\big(1 - \lambda(\theta)\big) \le T\lambda(\theta),
\qquad
\mathrm{Cov}[h(\theta), h(\theta')] = -T\lambda(\theta)\lambda(\theta') \le 0 \quad (\theta \ne \theta').
\]
Since $\hat c(\varphi)$ is a positively weighted sum of the $h(\theta)$, with weights $1/(T\lambda(\theta))$, the
covariance terms can only reduce its variance, and dropping them gives
\[
\mathrm{Var}[\hat c(\varphi)] \le \sum_{\theta \in \instances{\varphi}} \frac{T\lambda(\theta)}{T^2\lambda(\theta)^2}
= \sum_{\theta \in \instances{\varphi}} \frac{1}{T\lambda(\theta)} \le \frac{n}{T\lambda_{\min}},
\]
where $n = c(\varphi)$. Dividing by $c(\varphi)^2 = n^2$, and using $q(\varphi) = \sum_\theta \lambda(\theta) \le n\lambda_{\max}$
and~\eqref{eqn:samplingRatio},
\[
\frac{\mathrm{Var}[\hat c(\varphi)]}{c(\varphi)^2} \le \frac{1}{T n \lambda_{\min}} \le \frac{d_{\max}^{D}}{T n \lambda_{\max}}
\le \frac{d_{\max}^{D}}{T q(\varphi)}.
\]
By (A1), every pattern among the $M$ most frequent has $q(\varphi) \ge 1/(M H_l)$, so the relative variance is
at most $d_{\max}^{D} M H_l / T$. Hence $T \ge M H_l/\varepsilon'^2$ gives a relative variance of at most
$d_{\max}^{D}\varepsilon'^2$, i.e.\ a relative root-mean-square error of at most $d_{\max}^{D/2}\varepsilon'$, for every
top-$M$ pattern.

Algorithm~\ref{alg:fragmentMining} instead starts exactly $N$ paths at every node. This is stratified
sampling with the nodes as strata. Since each node has weight $1/|V|$ under a uniformly random start and
receives $N$ of the $T$ paths (proportional allocation), the variance of $\hat c(\varphi)$ is no larger than
under independent random starts \citep[][Ch.~5]{cochran1977sampling}.
\end{proof}

\begin{lemma}[Number of patterns]\label{lem:numberOfPatterns}
$\ln |\mathbf{G}_l| \le l \ln(2|\mathcal{P}|)$, and hence $H_l \le 1 + L_{\text{max}} \ln(2|\mathcal{P}|)$.
\end{lemma}

\begin{proof}
A pattern of length $l$ is a sequence of $l$ atoms. Each atom is one of the $|\mathcal{P}|$ predicates,
and each binary atom can be oriented in one of two directions along the pattern, so
$|\mathbf{G}_l| \le (2|\mathcal{P}|)^l$. The bound on $H_l$ follows from $H_n \le 1 + \ln n$ and $l \le L_{\text{max}}$.
\end{proof}

\begin{lemma}[Precision uncertainty]\label{lem:precision}
If $\hat c(\body{\Rule})$ and $\hat c(\body{\Rule} \wedge \head{\Rule})$ are $\varepsilon'$-uncertain,
then $\precisionest{\Rule}$ is $(2\varepsilon' + \mathcal{O}(\varepsilon'^2))$-uncertain.
\end{lemma}

\begin{proof}
$\precisionest{\Rule} = \hat c(\body{\Rule} \wedge \head{\Rule}) / \hat c(\body{\Rule})$. To first order
(the delta method), the relative error of a ratio is the difference of the relative errors of its
numerator and denominator, whose root-mean-square is at most $2\varepsilon'$ by the triangle inequality
(Minkowski's inequality). The second-order terms, which include the bias of the ratio, are
$\mathcal{O}(\varepsilon'^2)$.
\end{proof}

\begin{lemma}[Recall uncertainty]\label{lem:recall}
For a rule set $\Rules_\alpha$ with head $\alpha$, let $d(\logicfact) = \sum_{\Rule \in \Rules_\alpha}
|\restrictedinstances{\logicfact}|$ be the number of ground patterns entailing head fact $\logicfact$, and for a
ground pattern $\theta$ entailing $\logicfact_\theta$ let $w(\theta) = \ln(1 + d(\logicfact_\theta)) / d(\logicfact_\theta)$.
Then
\[
\recall{\Rules_\alpha} = \sum_{\logicfact} \ln\big(1 + d(\logicfact)\big) = \sum_{\theta} w(\theta),
\qquad \hat R(\Rules_\alpha) \vcentcolon= \sum_{\theta} \frac{h(\theta)\, w(\theta)}{T \lambda(\theta)}
\]
is unbiased. If the rule patterns of $\Rules_\alpha$ are among the $M$ most frequent patterns of their
length and $|V| N \ge M H/\varepsilon'^2$, where $H = \max_{l \le L_{\text{max}}} H_l$, then $\hat R(\Rules_\alpha)$ is $\sqrt{L_{\text{max}}}\, d_{\max}^{D}\varepsilon'$-uncertain.
\end{lemma}

\begin{proof}
The identity holds because each fact $\logicfact$ is entailed by exactly $d(\logicfact)$ ground patterns, each
contributing $\ln(1 + d(\logicfact))/d(\logicfact)$. 
Unbiasedness follows from $\mathbb{E}[h(\theta)] = T\lambda(\theta)$.

\emph{Variance.} The rules of $\Rules_\alpha$ may have different lengths. Write $\hat R = \sum_l \hat R_l$, where $\hat R_l$
sums over the ground patterns of length $l$. Within one length, a path traverses at most one ground pattern,
so as in the proof of Lemma~\ref{lem:counts} the counts are multinomial and
$\mathrm{Var}[\hat R_l] \le \sum_{\theta} w(\theta)^2/(T\lambda(\theta))$, summing over $\theta$ of length $l$. Across
lengths, one path can traverse a ground pattern of each length, so the $\hat R_l$ may be positively
correlated; since there are at most $L_{\text{max}}$ lengths, the Cauchy--Schwarz inequality gives
$\mathrm{Var}[\hat R] \le L_{\text{max}} \sum_l \mathrm{Var}[\hat R_l]$. Hence
\[
\mathrm{Var}[\hat R] \le L_{\text{max}} \sum_\theta \frac{w(\theta)^2}{T\lambda(\theta)} \le \frac{L_{\text{max}}}{T\lambda_{\min}} \sum_\theta w(\theta)^2
\le \frac{L_{\text{max}} \ln 2}{T\lambda_{\min}}\, \recall{\Rules_\alpha},
\]
where the last step uses $w(\theta) \le \ln 2$ (since $\ln(1 + x)/x$ is decreasing and $d \ge 1$), so
$\sum_\theta w(\theta)^2 \le \ln 2 \sum_\theta w(\theta) = \ln 2 \cdot \recall{\Rules_\alpha}$.

\emph{Relating $\lambda_{\min}$ to frequency.} Let $n = \sum_{\logicfact} d(\logicfact)$ be the number of ground patterns and
$q = \sum_\theta \lambda(\theta) \le n\lambda_{\max}$. By~\eqref{eqn:samplingRatio},
$\lambda_{\min} \ge \lambda_{\max} d_{\max}^{-D} \ge q\, d_{\max}^{-D}/n$. Substituting and dividing by $\recall{\Rules_\alpha}^2$ gives,
\[
\frac{\mathrm{Var}[\hat R]}{\recall{\Rules_\alpha}^2} \le \frac{L_{\text{max}} \ln 2 \cdot d_{\max}^{D}}{T q} \cdot \frac{n}{\recall{\Rules_\alpha}}.
\]

\emph{Bounding $n/\recall{\Rules_\alpha}$.} A rule's body is a path or cycle of at most $D$ binary edges starting
at a node of $\logicfact$, with its unary atoms fixed by the rule, so at most $d_{\max}^{D}$ ground patterns of one
rule entail $\logicfact$, and $d(\logicfact) \le |\Rules_\alpha|\, d_{\max}^{D}$. Since $x/\ln(1 + x)$ is increasing,
each fact has $d(\logicfact)/\ln(1 + d(\logicfact)) \le |\Rules_\alpha| d_{\max}^{D}/\ln 2$, and therefore
\[
\frac{n}{\recall{\Rules_\alpha}} = \frac{\sum_{\logicfact} d(\logicfact)}{\sum_{\logicfact} \ln(1 + d(\logicfact))} \le \frac{|\Rules_\alpha|\, d_{\max}^{D}}{\ln 2}.
\]

\emph{Bounding $Tq$.} $q$ sums the frequencies $q(\body{\Rule} \wedge \head{\Rule})$ of the $|\Rules_\alpha|$ rule
patterns. Each is at least $1/(M H_l) \ge 1/(M H)$ by (A1), so $T q \ge |\Rules_\alpha| T/(M H) \ge |\Rules_\alpha|/\varepsilon'^2$.

Combining the three bounds, the relative variance is at most $L\, d_{\max}^{2D}\varepsilon'^2$, i.e.\ the relative
root-mean-square error is at most $\sqrt{L_{\text{max}}}\, d_{\max}^{D}\varepsilon'$.
\end{proof}

\begin{manualtheorem}{\ref{theorem:N}}[Optimality]
Under (A1), let $\Rules$ be a set of rules whose body and rule patterns are among the $M$ most
frequent patterns of their length, with $M \ge 2|\mathcal{P}|$. If
\[
N = \mathcal{O}\!\left(\frac{M D \ln M}{\varepsilon^2 |V|}\right),
\]
then $\utilityE{\Rules}$ is $\varepsilon$-uncertain.
More precisely, $N = 9 L\, d_{\max}^{2D}\, M H / (\varepsilon^2|V|)$ suffices.
\end{manualtheorem}

\begin{proof}
Let $\varepsilon' = \varepsilon / (3 \sqrt{L_{\text{max}}}\, d_{\max}^{D})$ and $N = M H / (\varepsilon'^2 |V|)$. By Lemma~\ref{lem:counts}, every
required pattern count is $d_{\max}^{D/2}\varepsilon'$-uncertain, so by Lemma~\ref{lem:precision} each precision
estimate is $(2 d_{\max}^{D/2}\varepsilon' + \mathcal{O}(\varepsilon'^2))$-uncertain. By Lemma~\ref{lem:recall}, each rule-set
recall estimate is $\sqrt{L}\, d_{\max}^{D}\varepsilon'$-uncertain. The utility (Definition~\ref{definition:utilityRuleSet}) is a sum,
over head atoms $\alpha$, of products $A_\alpha B_\alpha$, where $A_\alpha$ is a positively weighted sum of
precisions and $B_\alpha = \recall{\Rules_\alpha}\complexity{\Rules_\alpha}$. A positively weighted sum of
estimates, each with relative error at most $\delta$, again has relative error at most $\delta$
(Minkowski's inequality), and to first order the relative error of a product is at most the sum of the
relative errors. Hence $\utilityE{\Rules}$ is $\big((2 d_{\max}^{D/2} + \sqrt{L_{\text{max}}}\, d_{\max}^{D})\varepsilon' + \mathcal{O}(\varepsilon'^2)\big)$-uncertain,
and $(2 d_{\max}^{D/2} + \sqrt{L_{\text{max}}}\, d_{\max}^{D})\varepsilon' \le 3 \sqrt{L_{\text{max}}}\, d_{\max}^{D}\varepsilon' = \varepsilon$. Finally, by
Lemma~\ref{lem:numberOfPatterns} and $2|\mathcal{P}| \le M$, $\ln|\mathbf{G}| \le L_{\text{max}}\ln(2|\mathcal{P}|) \le (2D + 1)\ln M$,
so $N = \mathcal{O}\big(M D \ln M / (\varepsilon^2|V|)\big)$.
\end{proof}

\begin{remark}[Choice of $N$ in practice]\label{rmk:N}
\label{sec:justRmkN}
The factor $D$  comes from the worst-case bound $\ln|\mathbf{G}| \le (2D+1)\ln M$, which assumes
that every sequence of predicates occurs as a pattern. In practice, far fewer patterns occur. To first approximation, it is thus reasonable to discount the $D$ dependence ($D=3$ is constant in our experiments). Hence SPECTRUM sets $N = \lceil M \ln M / (\varepsilon^2|V|) \rceil$: it uses $\ln M$ as an estimate of $\ln|\mathbf{G}|$, and
absorbs constant factors into $\varepsilon$. With the heuristic $M = 20|\mathcal{P}|$, this estimate is
accurate on our benchmarks: UMLS has $|\mathbf{G}| \approx 1{,}670$ mined patterns against $M = 880$, and
FB15K-237 has $|\mathbf{G}| \approx 15{,}800$ against $M = 4{,}740$, so $\ln M$ and $\ln|\mathbf{G}|$ differ by
less than 20\%.
\end{remark}

\subsection{Noise Robustness}
\label{section:proof:noise}

\paragraph{Noise model.} Each fact of $\trainingdata$ is independently deleted with probability $p$.
This models the incompleteness of real knowledge graphs. We compare quantities in expectation, writing
$\tilde{\ }$ for the noisy data; for precision, which is a ratio of counts, we take the ratio of the
expected counts, which is accurate to leading order when the counts are large.

\begin{manualtheorem}{\ref{thm:noise_robustness}}[Noise robustness]
Under
this noise model, for every rule set $\Rules$, the noisy utility $\tilde U(\Rules)$ is within a fractional
error $\mathcal{O}(Lp)$ of the true utility $U(\Rules)$. Consequently, if SPECTRUM is run on the noisy data
with $N$ as in Thm.~\ref{theorem:N}, its utility estimate $\hat U(\Rules)$ is within a fractional error
$\varepsilon + \mathcal{O}(Lp)$ of the true utility $U(\Rules)$.
\end{manualtheorem}

\begin{proof}
We follow the effect of deletion through each factor of the utility in turn.

\emph{Pattern counts.} A ground pattern of length $l$ consists of $l$ facts. It is still present in the
noisy data only if none of them is deleted, which happens with probability $(1 - p)^l$, independently of
which pattern it belongs to. Summing over the ground patterns of $\varphi$,
\[
\mathbb{E}[\tilde c(\varphi)] = (1 - p)^l\, c(\varphi).
\]

\emph{Precision.} The precision of $\Rule$ is the ratio $c(\body{\Rule} \wedge \head{\Rule}) / c(\body{\Rule})$,
multiplied by the symmetry factor and the prior. If the body has length $l$, the rule pattern has length
$l + 1$, so the ratio of expected counts is
\[
\frac{(1 - p)^{l+1}\, c(\body{\Rule} \wedge \head{\Rule})}{(1 - p)^{l}\, c(\body{\Rule})}
= (1 - p)\, \frac{c(\body{\Rule} \wedge \head{\Rule})}{c(\body{\Rule})}.
\]
Intuitively, a body grounding whose head fact has been deleted now counts against the rule. The Bayesian
prior is a ratio of predicate frequencies, all of which are rescaled by $1 - p$, so it is unchanged. The
symmetry depends only on the shape of the rule. Hence the precision is rescaled by
exactly $1 - p$.

\emph{Recall.} Fix a head fact $\logicfact$ with degree $d = d(\logicfact) \ge 1$. The fact contributes to the noisy
recall only if it survives, which happens with probability $1 - p$. Given that it survives, each of its
$d$ ground patterns survives if its remaining facts do. A ground pattern has at most $L_{\text{max}}$ facts, so this
happens with probability at least $(1 - p)^{L_{\text{max}}-1}$, and the noisy degree $\tilde d$ satisfies
$\mathbb{E}[\tilde d] \ge (1 - p)^{L_{\text{max}}-1} d$. We bound $\mathbb{E}[\ln(1 + \tilde d)]$ from both sides:
\begin{itemize}[leftmargin=*]
\item \emph{Upper bound.} $\tilde d \le d$ and $\ln(1 + x)$ is increasing, so $\mathbb{E}[\ln(1 + \tilde d)] \le \ln(1 + d)$.
\item \emph{Lower bound.} $\ln(1 + x)$ is concave, so on $0 \le x \le d$ it lies above the straight line
through its endpoints, $\ln(1 + x) \ge \frac{x}{d}\ln(1 + d)$. Taking expectations,
$\mathbb{E}[\ln(1 + \tilde d)] \ge \frac{\mathbb{E}[\tilde d]}{d}\ln(1 + d) \ge (1 - p)^{L_{\text{max}}-1}\ln(1 + d)$.
\end{itemize}
Multiplying by the survival probability $1 - p$ of the head fact, its expected contribution lies
between $(1 - p)^{L_{\text{max}}}\ln(1 + d)$ and $(1 - p)\ln(1 + d)$. Summing over the head facts gives
\[
(1 - p)^{L_{\text{max}}}\, \recall{\Rules_\alpha} \;\le\; \mathbb{E}[\tilde R(\Rules_\alpha)] \;\le\; (1 - p)\,\recall{\Rules_\alpha}.
\]
The complexity factors only depend on the rule shape, so are not affected.

\emph{Utility.} The utility is a positively weighted sum, over head atoms $\alpha$, of products $A_\alpha B_\alpha$,
where $A_\alpha$ is a positively weighted sum of precisions and $B_\alpha = \recall{\Rules_\alpha}\complexity{\Rules_\alpha}$.
By the steps above, $A_\alpha$ is rescaled by $1 - p$ and $B_\alpha$ by a factor between $(1 - p)^{L_{\text{max}}}$ and $1 - p$,
so each product, and hence the utility, is rescaled by a factor between $(1 - p)^{L_{\text{max}}+1}$ and $(1 - p)^2 \le 1$:
\[
(1 - p)^{L+1}\, \utility{\Rules} \;\le\; \tilde U(\Rules) \;\le\; \utility{\Rules}.
\]
Therefore $0 \le \utility{\Rules} - \tilde U(\Rules) \le \big(1 - (1 - p)^{L_{\text{max}}+1}\big)\utility{\Rules} \le (L_{\text{max}} + 1)\,p\,\utility{\Rules}$,
where the last step is Bernoulli's inequality $(1 - p)^{L_{\text{max}}+1} \ge 1 - (L_{\text{max}} + 1)p$. Dividing by $\utility{\Rules}$ gives the first claim.

\emph{SPECTRUM's estimates.} SPECTRUM only ever sees the noisy data, so Thm.~\ref{theorem:N}, applied to
the noisy data (with $N$ computed from the noisy graph), makes $\hat U(\Rules)$ an $\varepsilon$-uncertain
estimate of $\tilde U(\Rules)$:
$\sqrt{\mathbb{E}\big[(\hat U(\Rules) - \tilde U(\Rules))^2\big]} \le \varepsilon\, \tilde U(\Rules) \le \varepsilon\, \utility{\Rules}$,
where the last step uses $\tilde U(\Rules) \le \utility{\Rules}$. By the triangle inequality for the
root-mean-square error and the bound above,
\[
\sqrt{\mathbb{E}\big[(\hat U(\Rules) - \utility{\Rules})^2\big]}
\;\le\; \sqrt{\mathbb{E}\big[(\hat U(\Rules) - \tilde U(\Rules))^2\big]} + \big(\utility{\Rules} - \tilde U(\Rules)\big)
\;\le\; \big(\varepsilon + (L_{\text{max}} + 1)p\big)\,\utility{\Rules},
\]
so $\hat U(\Rules)$ is $(\varepsilon + (L+1)p)$-uncertain as an estimate of the true utility.
\end{proof}

\subsection{Complexity of the Utility Metric}
\label{appx:utilityComplexity}

\begin{theorem}\label{thm:utilityComplexity}
Represent a relational database as a hypergraph with edge set $E$ and $|\mathcal{P}|$ predicates.
Suppose that, for every rule $\Rule$ of length at most $D$ and every head grounding $\logicfact$, the
sets $\instances{\body{\Rule}}$, $\instances{\head{\Rule}}$, $\instances{\body{\Rule} \wedge \head{\Rule}}$
and $\restrictedinstances{\logicfact}$ have been computed during pattern mining. Then computing the
utility of a rule takes $\mathcal{O}(|E| + D + |\mathcal{P}|)$ time, and that of a rule set $\Rules$ takes
$\mathcal{O}(|\Rules|(|E| + D + |\mathcal{P}|))$ time.
\end{theorem}

\begin{proof}
We bound the cost of each factor of the utility for a single rule.
\emph{Precision and complexity factor}: a ratio of two precomputed set sizes and the rule's length,
$\mathcal{O}(1)$.
\emph{Symmetry factor}: precomputed for every pattern shape of length at most $D$; identifying the
rule's shape takes one pass over its atoms, $\mathcal{O}(D)$.
\emph{Bayesian prior}: its denominator sums over the predicates with the same argument types as
$\head{\Rule}$, $\mathcal{O}(|\mathcal{P}|)$.
\emph{Recall}: one pass over the head groundings, at most $\mathcal{O}(|E|)$.
Summing gives $\mathcal{O}(|E| + D + |\mathcal{P}|)$. For a rule set, the recall of each partition
$\Rules_\alpha$ costs $\mathcal{O}(|\Rules_\alpha||E|)$ and the remaining factors
$\mathcal{O}(|\Rules_\alpha|(D + |\mathcal{P}|))$; summing over the partitions gives
$\mathcal{O}(|\Rules|(|E| + D + |\mathcal{P}|))$.
Since the greedy selection of Alg.~\ref{alg:pipeline} returns at most $M$ rules and evaluates a
rule-set utility $\mathcal{O}(M^2)$ times, the selection loop costs $\mathcal{O}(M^3(|E| + D + |\mathcal{P}|))$
in the worst case.
\end{proof}

\section{Experimental Details}\label{app:experimentDetails}

\subsection{Setup}

\paragraph{Hardware.} SPECTRUM runs on a 6-core 2.60\,GHz i7-10750H CPU. All baselines run on a
V100 GPU (32\,GB memory, 40 CPUs) with the hyperparameters recommended by their authors.

\paragraph{Hyperparameters.} Unless stated otherwise, we use $D = 3$, $\varepsilon = 0.01$ and
$M = 20|\mathcal{P}|$, where $|\mathcal{P}|$ is the number of predicates in the dataset. $M$ and
$\varepsilon$ set the number of paths sampled per node, $N$ (Section~\ref{section:pattern mining});
App.~\ref{appx:sensitivityEpsilonAndM} shows that results are insensitive to both. 

\paragraph{Evaluation.} For knowledge graph completion we implemented a C++ evaluation script, following the methodology of NCRL
\citep{cheng2023neural}, but with $\sim 100\times$ faster execution time. Each rule is weighted by its precision (corrected for symmetry and prior,
Section~\ref{section:utility}), and each candidate entity is scored by the sum of the weights of the
rules that predict it. We report mean reciprocal rank (MRR) and Hit@$k$.


\subsection{Knowledge Graph Completion}
\label{sec:appx:KGC}

\paragraph{Datasets.} Table~\ref{tab:datasets} summarises the five benchmark datasets.

\begin{table}[h!]
\caption{Training-set statistics of the knowledge graph completion benchmarks.}
\label{tab:datasets}
\centering
\begin{tabular}{lrr}
\toprule
Dataset   & Facts   & Predicates \\ \midrule
Family    & 5,868   & 12  \\
UMLS      & 1,302   & 41  \\
Kinship   & 2,350   & 24  \\
WN18RR    & 86,835  & 11  \\
FB15K-237 & 272,115 & 237 \\
\bottomrule
\end{tabular}
\end{table}

\begin{figure}[h!]
\centering
\includegraphics[width=0.6\linewidth]{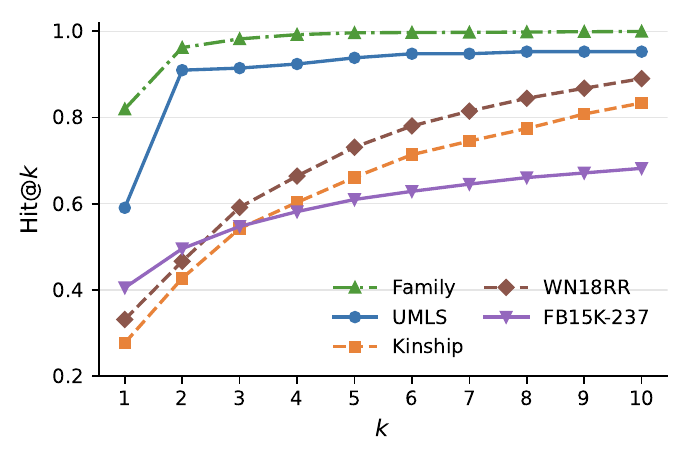}
\caption{Hit@$k$ of SPECTRUM for $k = 1, \dots, 10$ on the five knowledge graph completion
benchmarks.}
\label{fig:hitsAtK}
\end{figure}

\paragraph{Hit@$k$.} Figure~\ref{fig:hitsAtK} extends the Hit@10 results of
Table~\ref{tab:combined_metrics_runtimes} to all $k$ from 1 to 10.

\paragraph{Additional baselines.} We considered several further structure learners. They were
excluded because they are not competitive, solve a different task, or do not scale to our
benchmarks.
\begin{itemize}[leftmargin=*]
\item \textbf{RLogic} \citep{cheng2022rlogic} reports MRR 0.47 and Hit@10 0.537 on WN18RR, and
MRR 0.31 and Hit@10 0.501 on FB15K-237. No evaluation script is available, so we could not
reproduce these results. It is much slower than SPECTRUM: 170\,s on Family (SPECTRUM: 0.2\,s),
180\,s on WN18RR (0.9\,s), and over 24 hours on FB15K-237 (17\,s).
\item \textbf{IterE} \citep{zhang2019iteratively} reports MRR 0.27 and Hit@10 0.31 on WN18RR, and
MRR 0.24 and Hit@10 0.39 on FB15K-237, well below SPECTRUM and NCRL
(Table~\ref{tab:combined_metrics_runtimes}).
\item \textbf{Neural LP} \citep{NeuralLP} is outperformed by NCRL on all our benchmarks
\citep{cheng2023neural}.
\item \textbf{ProbFOIL} \citep{deraedt2015inducing} learns only chain rules over binary predicates,
$P_1(X, Z_1) \land \dots \land P_{n-1}(Z_{n-1}, Y) \rightarrow P_n(X, Y)$, since each new atom may
introduce at most one new variable. SPECTRUM also learns non-chain rules, which are essential for
data with unary predicates, e.g., SPECTRUM learns the following non-chain rules in our benchmark datasets:
    \begin{itemize}
        \item IMDB: $\textsc{actor}(X) \land \textsc{workedUnder}(X, Y) \land \textsc{workedUnder}(Z, Y) \rightarrow \textsc{actor}(Z)$
        \item CAD: $\textsc{queues}(X) \land \textsc{closeTo}(X, Y) \rightarrow \textsc{queues}(Y)$
    \end{itemize}
\item \textbf{R5} \citep{lu2022r5} is trained on queries $(G, (s, X, o), r)$, asking for the
relation $r$ between entities $s$ and $o$ in a graph $G$, rather than on the facts of a single
database.
\item \textbf{Neural Logic Machines} \citep{NLMs} and \textbf{Differentiable Logic Machines}
\citep{zimmer2023differentiable} learn a supervised mapping from input to target graphs, rather
than completing a single graph. They are evaluated on synthetic datasets of at most 350 entities, and run out of memory beyond this at depth 3. In contrast, SPECTRUM scales linearly to real-world datasets with $10^4$-$10^6$ entities. 
\item \textbf{$\alpha$ILP} \citep{shindo2023alphailp} learns from images of positive and negative
scenes, about 20 scenes of a few dozen objects each, rather than from relational facts without
negative examples. A comparison would require substantially re-engineering it.
\end{itemize}

\subsection{PSL Experiments}
\label{appx:PSL}

\paragraph{Datasets.} Table~\ref{tab:psl_datasets} summarises the two datasets.
\begin{itemize}[leftmargin=*]
\item \textbf{CAD} (collective activity detection) describes people, drawn as bounding boxes, and
the actions they perform (crossing, waiting, queuing, walking, talking, dancing, jogging) over a
sequence of video frames. $\textsc{frame}(B, F)$ states that box $B$ is in frame $F$;
$\textsc{flabel}(F, a)$ that most people in frame $F$ perform action $a$; $\textsc{doing}(B, a)$
that the person in box $B$ performs action $a$; $\textsc{close}(B_1, B_2)$ that two boxes in a frame
are close; and $\textsc{same}(B_1, B_2)$ that two boxes in different frames show the same person.
\item \textbf{Yelp} (2020) contains users' ratings of local businesses, business categories and
friendships between users. Using the preprocessing of Kouki et al.~\cite{kouki2015hyper}, we obtain the
predicates $\textsc{rating}$, $\textsc{friends}$, $\textsc{similarUsers}$, $\textsc{similarItems}$,
$\textsc{averageUserRating}$ and $\textsc{averageItemRating}$.
\end{itemize}

\begin{table}[h!]
\caption{Training-set statistics of the PSL datasets.}
\label{tab:psl_datasets}
\centering
\begin{tabular}{lrr}
\toprule
Dataset & Facts                 & Predicates \\ \midrule
CAD     & $2.4 \times 10^5$ & 19 \\
Yelp    & $2.2 \times 10^6$ & 15 \\
\bottomrule
\end{tabular}
\end{table}

\paragraph{Recovering hand-crafted rules.} We compare the rules SPECTRUM learns with those written
by domain experts.

\emph{CAD.} To learn separate rules per action, we replace $\textsc{doing}(B, a)$ with one unary
predicate $\textsc{doing}_a(B)$ per action $a$. SPECTRUM then recovers all 21 hand-crafted rules of London et al.~\cite{collective-psl}: for each of the seven actions $a$,
\begin{align*}
&\textsc{frame}(B, F) \land \textsc{flabel}(F, a) \rightarrow \textsc{doing}_a(B), \\
&\textsc{doing}_a(B_1) \land \textsc{close}(B_1, B_2) \rightarrow \textsc{doing}_a(B_2), \\
&\textsc{doing}_a(B_1) \land \textsc{same}(B_1, B_2) \rightarrow \textsc{doing}_a(B_2).
\end{align*}

\emph{Yelp.} To see whether high and low ratings propagate differently, we split $\textsc{rating}$
into $\textsc{rating}_{\text{high}}$ and $\textsc{rating}_{\text{low}}$. For $x \in
\{\text{high}, \text{low}\}$, SPECTRUM learns
\begin{align*}
&\textsc{rating}_x(U_1, I) \land \textsc{similarUsers}(U_1, U_2) \rightarrow \textsc{rating}_x(U_2, I), \\
&\textsc{rating}_x(U, I_1) \land \textsc{similarItems}(I_1, I_2) \rightarrow \textsc{rating}_x(U, I_2), \\
&\textsc{rating}_x(U_1, I) \land \textsc{friends}(U_1, U_2) \rightarrow \textsc{rating}_x(U_2, I),
\end{align*}
with each similarity predicate instantiated for four measures (Pearson, cosine, latent cosine and
latent Euclidean). This recovers 24 of the 26 hand-crafted rules of Kouki et al.~\cite{kouki2015hyper}.
The two it cannot learn,
$\textsc{averageItemRating}(I) \leftrightarrow \textsc{rating}(U, I)$ and
$\textsc{averageUserRating}(U) \leftrightarrow \textsc{rating}(U, I)$, are not term-constrained
(Section~\ref{section:theory}), since $U$ and $I$ respectively do not appear in the body.

\section{Ablation Studies}
\label{appx:ablations}

\subsection{Assessing SPECTRUM's Sensitivity to $\varepsilon$ and $M$}
\label{appx:sensitivityEpsilonAndM}

SPECTRUM has two main hyperparameters, $M$ and $\varepsilon$. Together they set the number of paths
$N$ sampled from each node (Section~\ref{section:pattern mining}): a larger $M$ or a smaller
$\varepsilon$ gives a larger $N$, and hence more complete mining at a higher cost. In our main
experiments we use $\varepsilon = 0.01$ and a heuristic $M = 20 \times$ (number of predicates in the
dataset). Here we test how sensitive SPECTRUM's results are to these choices.

\paragraph{Method.} We vary $M$ from $2$ to $2^{17}$ with $\varepsilon = 0.01$ fixed, and
$\varepsilon$ from $0.1$ to $0.001$ with $M$ fixed at the heuristic value. For each setting, we record
the number of rules found and the runtime, over three repeats.

\begin{figure}
    \centering
    \includegraphics[width=\linewidth]{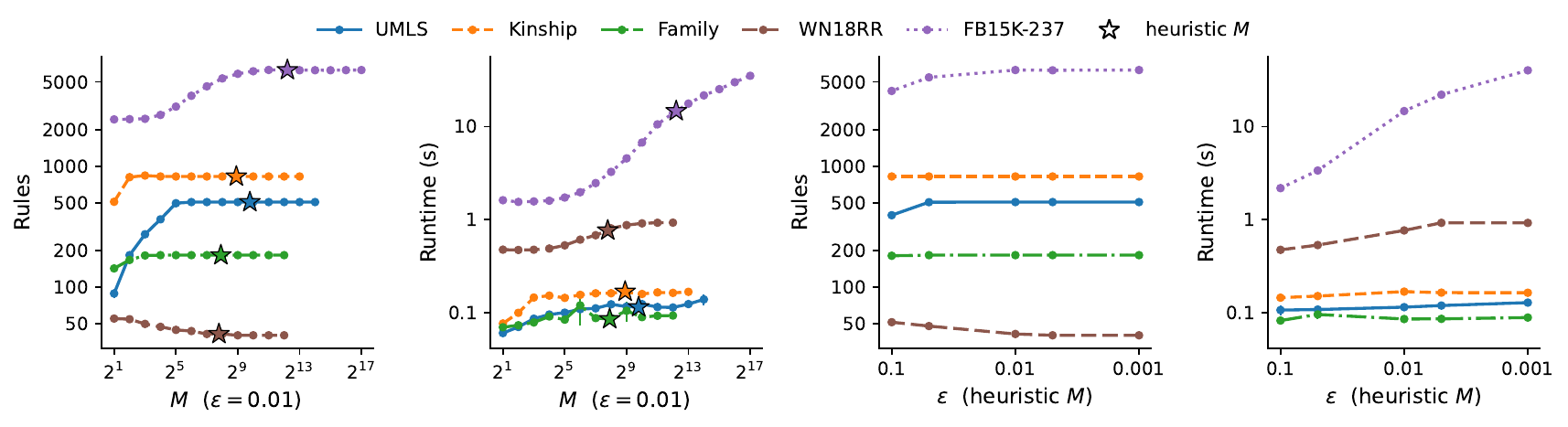}
    \caption{Rules found and runtime of SPECTRUM as $M$ varies with $\varepsilon = 0.01$ fixed
    (left two panels), and as $\varepsilon$ varies with the heuristic $M$ fixed (right two panels).
    In every panel the sampling budget grows to the right. Stars mark the heuristic
    $M = 20 \times$ (number of predicates). Mean $\pm$ standard deviation over 3 runs; error bars are
    mostly smaller than the markers.}
    \label{fig:ablationStudy}
\end{figure}

\paragraph{Results.} Figure~\ref{fig:ablationStudy} shows the results. As $M$ grows, the number of
rules rises and then levels off. This happens well below the heuristic $M$ on every dataset: by
$M = 8$ on Family and Kinship, $M = 64$ on UMLS and $M = 2048$ on FB15K-237. Beyond this point,
increasing $M$ costs runtime without finding more rules. On FB15K-237, runtime grows from 15\,s at the
heuristic $M$ to 35\,s at $M = 2^{17}$. On the three smallest datasets, runtime stays below 0.2\,s
throughout, because SPECTRUM already explores every node's neighbourhood completely. On WN18RR, the
number of rules \emph{falls} as $M$ grows. As in Section~\ref{section:rw_comparison}, this is an
artefact of small budgets: with few paths per node, precision is overestimated and spurious rules
pass the precision threshold. SPECTRUM is similarly robust to $\varepsilon$: every dataset finds the
same rules for all $\varepsilon \le 0.01$. Only at $\varepsilon = 0.1$ do UMLS and FB15K-237 lose rules
(22\% and 33\% respectively).

\paragraph{Takeaway.} SPECTRUM is insensitive to $M$ and $\varepsilon$ over a wide range. The default
$\varepsilon = 0.01$ and heuristic $M$ find all the rules that larger budgets find, at close to
minimal runtime. 

\subsection{Scaling with Dataset Size}
\label{appx:scaling}

Figure~\ref{fig:scaling} compares SPECTRUM's runtime across datasets of different sizes. However,
these datasets also differ in structure: in density, number of predicates and how many patterns
each node takes part in. A fit across them therefore mixes the effect of size with the effect of
structure. Here we isolate the effect of size.

\paragraph{Method.} We enlarge each dataset by replication: the dataset with $k$ copies consists of
$k$ disjoint copies of the original graph, for $k \in \{1, 2, 4, 8, 16\}$. This multiplies the
number of nodes and facts by $k$ while leaving every node's neighbourhood unchanged. Since
$N = \lceil M \ln M / (\varepsilon^2 |V|) \rceil$ decreases as $|V|$ grows, we run two series:
\begin{itemize}[leftmargin=*]
\item \emph{Fixed $N$:} we set $\varepsilon_k = \varepsilon / \sqrt{k}$, which keeps the number of
paths per node constant. This tests the complexity result of Section~\ref{section:theory}, which
predicts that runtime grows linearly with the number of nodes.
\item \emph{Fixed $\varepsilon$:} we keep the default $\varepsilon$, as a user would, so that $N$
shrinks as $1/k$.
\end{itemize}
We use Family, UMLS, Kinship and WN18RR with their default hyperparameters, a single thread (to
exclude parallelisation effects) and three repeats.

\begin{figure}[tb]
\centering
\includegraphics[width=\linewidth]{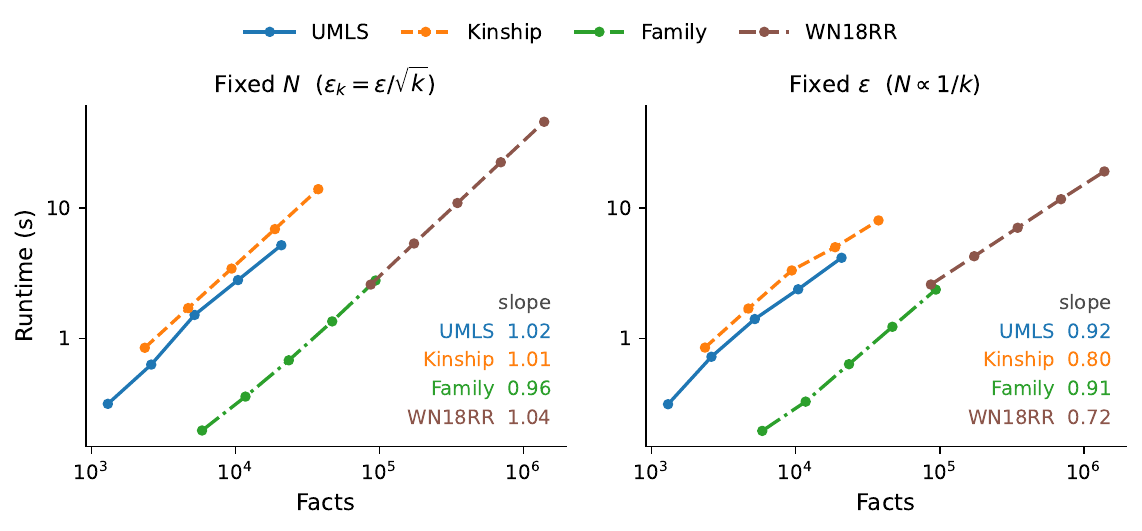}
\caption{SPECTRUM's runtime on datasets enlarged by replication ($k = 1, 2, 4, 8, 16$ disjoint
copies). Left: fixed paths per node ($\varepsilon_k = \varepsilon/\sqrt{k}$). Right: fixed
$\varepsilon$.}
\label{fig:replication}
\end{figure}

\paragraph{Results.} Figure~\ref{fig:replication} shows the results. With $N$ fixed (left), runtime
grows linearly with dataset size: the fitted log-log slopes are $1.02 \pm 0.04$ (UMLS),
$1.01 \pm 0.01$ (Kinship), $0.96 \pm 0.02$ (Family) and $1.04 \pm 0.01$ (WN18RR). With $\varepsilon$
fixed (right), runtime grows more slowly, with slopes from $0.72$ (WN18RR) to $0.92$ (UMLS). On
Family, UMLS and Kinship, $N$ is so large that SPECTRUM explores every neighbourhood completely, so
halving $N$ barely impacts runtime. Runtime is proportional to the number
of pattern occurrences found during mining, which is the quantity the
complexity analysis bounds.

\paragraph{Takeaway.} For a fixed number of paths per node, SPECTRUM's runtime is linear in dataset
size, as the complexity analysis predicts. 

\subsection{Comparing SPECTRUM's Pattern Mining to Random Walks}\label{section:rw_comparison}

A simpler way to mine patterns is to run $N$ random walks of length $D$ from
every node. Algorithm~\ref{alg:fragmentMining} improves on this in two ways
(Section~\ref{section:theory}): 
\begin{enumerate}
    \item It never wastes a step: random walks can backtrack along
edges they have already used, whereas Algorithm~\ref{alg:fragmentMining} only ever extends to an unvisited edge;
    \item Algorithm~\ref{alg:fragmentMining} is guaranteed to find every pattern within the $N$-close
neighbourhood of a node, and finds patterns beyond it with higher probability than random walks
(Theorem~\ref{theorem:N-closeness-completness}).
\end{enumerate}

Property 1 results in a difference in computational complexity, which we illustrate with an example: 

\begin{example}[Tree Sampling]
    Consider a depth $d$ tree with branching factor $b$, thus having $b^d$ distinct paths (or ground patterns). By the coupon collector's problem, finding all ground patterns requires an expected number $\mathbb{E}[N] = b^d H_{b^d} \approx b^d (\ln{b^d} + \gamma)$ of random walks, where $H_{b^d}$ is the $b^d$-th harmonic number and $\gamma$ is the Euler-Mascheroni constant. The ratio of the cost of sampling with random walks to the cost of Algorithm \ref{alg:fragmentMining} of discovering all these patterns is thus $O\left(\frac{b^d(\ln{b^d} + \gamma)}{b^d}\right) = O(d \ln{b})$. This proves that random walks have strictly worse computational complexity than Alg.~\ref{alg:fragmentMining} for this graph topology. 
\end{example}

We can also demonstrate the advantages of SPECTRUM's pattern-mining algorithm over random walks empirically:

\paragraph{Method.} We replace Algorithm~\ref{alg:fragmentMining} in the SPECTRUM pipeline with
random walks: from each node, $N$ independent walks of length up to $D$ are run. The rest of the pipeline is unchanged. Both variants use the same $N$ (set via
$\varepsilon$; Section~\ref{section:pattern mining}).
We vary $\varepsilon \in \{0.005, 0.01, 0.05, 0.1, 0.5\}$ and record the number of rules found and
the end-to-end runtime over three repeats.

\begin{figure}
    \centering
    \includegraphics[width=1.0\linewidth]{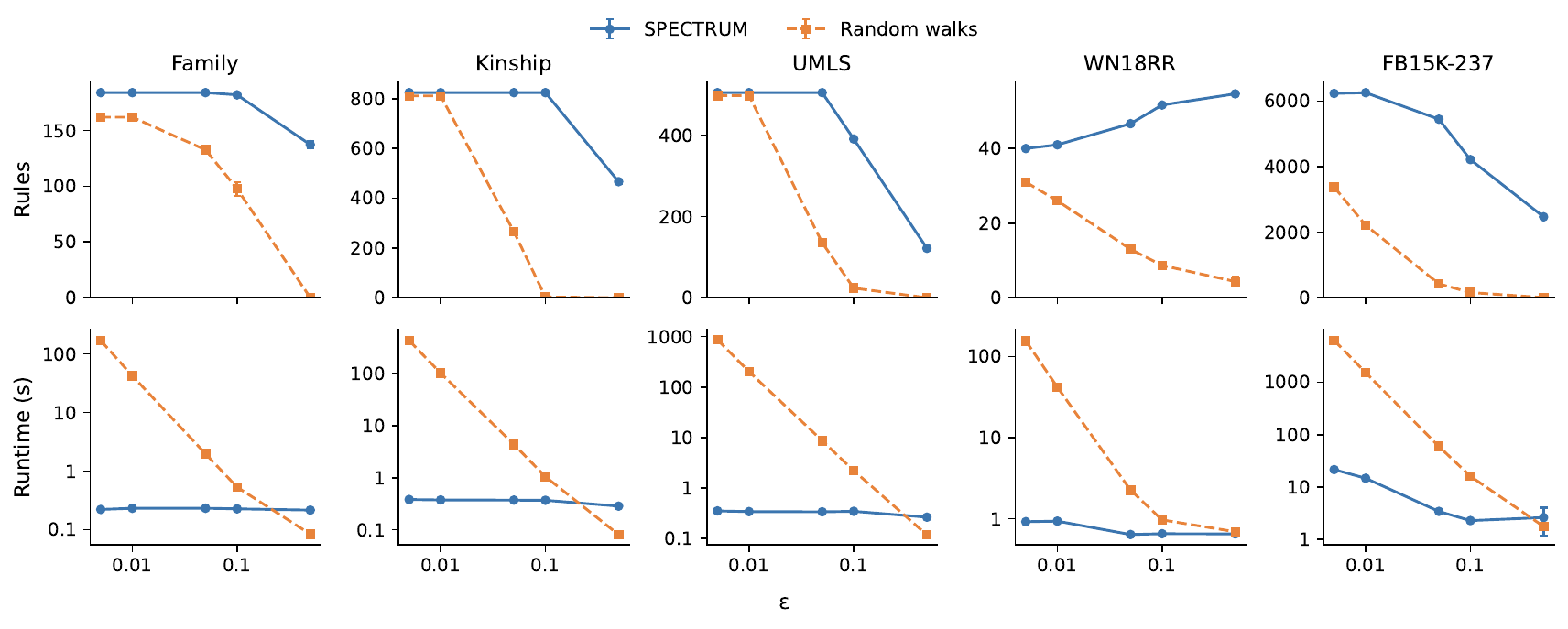}
    \caption{SPECTRUM's pattern mining (blue) against random walks (orange) as $\varepsilon$
    varies: rules found (top) and end-to-end runtime (bottom).}
    \label{fig:randomWalkAblation}
\end{figure}

\paragraph{Results.} Figure~\ref{fig:randomWalkAblation} shows the results. At every $\varepsilon$
and on every dataset, random walks find no more rules than SPECTRUM, and usually far fewer. At the
default $\varepsilon = 0.01$, they find 88\% (Family), 98\% (Kinship), 99\% (UMLS), 63\% (WN18RR)
and 35\% (FB15K-237) of the rules SPECTRUM finds, while taking anywhere from 45 to 600 times longer. Random walks are faster
only at $\varepsilon = 0.5$, where they find almost no rules. SPECTRUM's runtime, by contrast, is
nearly constant on the small datasets, because it already explores every node's neighbourhood
completely. On WN18RR, SPECTRUM appears to find more rules as $\varepsilon$ grows. This is an artefact of the
very small budget ($N = 1$ path per node at $\varepsilon = 0.5$): precision is then estimated from
very few sampled occurrences, and 16 of the 53 rules found at $\varepsilon = 0.5$ pass the
minimum-precision threshold only because their precision is overestimated. 

\paragraph{Takeaway.} For the same budget, SPECTRUM's pattern mining finds more rules than random
walks, and is one to three orders of magnitude faster.

\subsection{Assessing SPECTRUM's Robustness to Low-Recall, High-Precision Rules}

A useful rule should be both \emph{precise} (usually true when its body holds) and have high
\emph{recall} (it explains many facts; Section~\ref{section:utility}). However, to keep pattern mining fast,
SPECTRUM does not enumerate every pattern in the data. Could SPECTRUM therefore miss rules that are highly precise but
have low recall?

\paragraph{Method.} For each dataset, we first run SPECTRUM with an extremely large budget
($N = 10^9$ paths per node), which explores every node's neighbourhood completely. This yields every
rule SPECTRUM can learn. We then run SPECTRUM three times with its default hyperparameters and record
which of these rules (if any) it misses.

\begin{figure}
    \centering
    \includegraphics[width=1.0\linewidth]{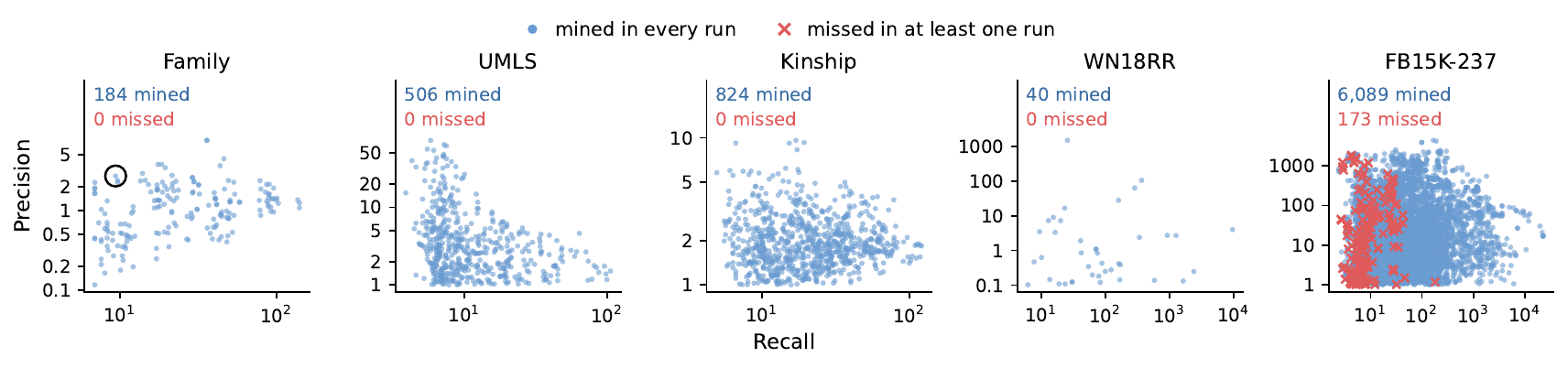}
    \caption{Precision (corrected for symmetry and prior) against recall for every rule SPECTRUM can
learn with precision above 1. Blue dots: rules found in all three runs with default
hyperparameters. Red crosses: rules missed in at least one run.}
    \label{fig:precisionRecallAblation}
\end{figure}

\paragraph{Results.} Figure~\ref{fig:precisionRecallAblation} plots precision against recall for
every learnable rule. Precise rules with low recall are \textit{common}: on every dataset, many of the most
precise rules lie at the low-recall end. For example, the Family rule ``if A and C are the mother
and father of the same child B, then A is the wife of C'' (circled) is in the 91st percentile by
precision but only the 8th by recall. Crucially, despite this fact, it is still found in every run, as are \textit{all other}
learnable rules on Family, UMLS, Kinship and WN18RR. Only on FB15K-237, the largest dataset, are
some rules missed: 173 of 6,262 (2.8\%) in at least one run. As
expected, these are low-recall rules: the miss rate is 6.3\% for the 25\% of rules with
the lowest recall and 0.1\% for the 25\% of rules with the highest.

\paragraph{Takeaway.} Although precise, low-recall rules are common, sampling rarely causes SPECTRUM
to miss them. It misses none on four of five datasets, and only $\sim2\%$ per run on FB15K-237.

\section{Pattern Mining Example}\label{section:pattern_mining_example}

\begin{example}\label{example:patternmining}
    We illustrate how Algorithm~\ref{alg:fragmentMining} mines patterns from the graph $\datagraph$ shown in Figure~\ref{fig:patternmining}. We
    follow a recursive call from node $v_0$ with parameters $N=4$ and $D=2$. To ease the presentation, we denote ground patterns as sets of edges. 
    
    Since $e_0$ is the only unary edge of $v_0$, Algorithm~\ref{alg:fragmentMining} stores the pattern 
    $\{e_0\}$ in $\globalpatterns$. Algorithm~\ref{alg:fragmentMining} then finds two binary edges $e_1$ and $e_2$ and, since $2 \leq N$, it considers both edges in turn. Algorithm~\ref{alg:fragmentMining} then grafts these two edges onto the previous pattern which by default contains the empty pattern $\emptyset$. In particular, for $e_1$, we graft $\emptyset \circ e_1 = \{e_1\}$ and $\{e_0\} \circ e_1  = \{e_0, e_1\}$. The resulting patterns along with $\previousedges =\{e_1\}$ are passed as 
    \[\finalpatterns = 
    \begin{Bmatrix} \{e_1\}\\\{e_0, e_1\}\end{Bmatrix}\] 
    to the next recursive call and are stored in $\globalpatterns$. 
    Algorithm~\ref{alg:fragmentMining} then proceeds analogously along $e_2$. After visiting node $v_2$, $\globalpatterns$ has as follows:
    \begin{align*}
        \globalpatterns = \begin{Bmatrix}
            \substack{\{e_0\}}\\
            \substack{\{e_1\}\\\{e_2\}}\\
            \substack{\{e_0, e_1\}\\\{e_0, e_2\}}\\
        \end{Bmatrix}.
    \end{align*}
    
    Both new recursions are started with updated $n=4/2=2$ and $d=1$. At $v_1$, since $n=2$ and since there are two, previously unvisited, edges $e_4$ and $e_5$, the algorithm continues the recursion along $e_4$ and $e_5$ with updated $n=2/2=1$ and $d=2$. In contrast, at $v_2$, since $n=2$ but there being three, previously unvisited, edges $e_7$, $e_8$, and $e_9$, the algorithm will choose at random two out of the three edges and continue the recursion with $n=2/2=1$. Since, $d=2$ in the next recursive calls, the algorithm terminates. Notice that if we set $N \geq 6$, then Algorithm~\ref{alg:fragmentMining} would find all ground patterns $\datagraph$. 
\end{example}

\begin{figure}[h]
    \centering
        \begin{tikzpicture}
	\begin{pgfonlayer}{nodelayer}
		\node [style=smallCircle] (0) at (0, 0) {$v_0$};
		\node [style=smallCircle] (1) at (-2, 0) {$v_1$};
		\node [style=smallCircle] (2) at (2, 0) {$v_2$};
		\node [style=smallCircle] (4) at (-3.5, 1) {$v_5$};
		\node [style=smallCircle] (5) at (-3.5, -1) {$v_4$};
		\node [style=smallCircle] (9) at (3.5, 1) {$v_8$};
		\node [style=smallCircle] (10) at (4, 0) {$v_7$};
		\node [style=smallCircle] (11) at (3.5, -1) {$v_6$};
		\node [style=none] (33) at (0, 0.75) {$e_0$};
		\node [style=none] (34) at (-1, 0.25) {$e_1$};
		\node [style=none] (35) at (1, 0.25) {$e_2$};
		\node [style=none] (36) at (-2, 0.75) {$e_3$};
		\node [style=none] (37) at (-3, 0.375) {$e_5$};
		\node [style=none] (38) at (-2.675, -0.75) {$e_4$};
		\node [style=none] (39) at (-3.5, -0.25) {$e_6$};
		\node [style=none] (40) at (2.675, 0.75) {$e_9$};
		\node [style=none] (41) at (2.675, -0.75) {$e_7$};
		\node [style=none] (42) at (3.5, -0.25) {$e_{10}$};
		\node [style=none] (43) at (3.25, 0.25) {$e_8$};
		\node [style=none] (44) at (4, 0.75) {$e_{11}$};
		\node [style=none] (45) at (5, 0.5) {};
		\node [style=none] (46) at (5.5, 0.5) {};
		\node [style=none] (47) at (5, 0) {};
		\node [style=none] (48) at (5.5, 0) {};
		\node [style=none] (49) at (5, -0.5) {};
		\node [style=none] (50) at (5.5, -0.5) {};
		\node [style=none] (51) at (6, 0.5) {$\textsc{p}_1$};
		\node [style=none] (52) at (6, 0) {$\textsc{p}_2$};
		\node [style=none] (53) at (6, -0.5) {$\textsc{p}_3$};
	\end{pgfonlayer}
	\begin{pgfonlayer}{edgelayer}
		\draw [style=blackdashed] (0) to (2);
		\draw [style=blackdashed] (0) to (1);
		\draw [style=undirected red, in=135, out=45, loop] (0) to ();
		\draw [style=undirected red, in=135, out=45, loop] (1) to ();
		\draw [style=undirected] (2) to (9);
		\draw [style=undirected] (11) to (2);
		\draw [style=undirected] (2) to (10);
		\draw [style=undirected] (1) to (5);
		\draw [style=undirected] (1) to (4);
		\draw [style=undirected red, in=135, out=45, loop] (11) to ();
		\draw [style=undirected red, in=135, out=45, loop] (10) to ();
		\draw [style=undirected red, in=135, out=45, loop] (5) to ();
		\draw [style=undirected red] (45.center) to (46.center);
		\draw [style=blackdashed] (47.center) to (48.center);
		\draw [style=undirected] (49.center) to (50.center);
	\end{pgfonlayer}
\end{tikzpicture}
        \caption[Pattern mining]{Graph $\datagraph$ from Example~\ref{example:patternmining}.
        The graph contains three types of labelled edges: red unary edges $\textsc{p}_1$, dashed black binary edges $\textsc{p}_2$, and solid black binary edges $\textsc{p}_3$.} \label{fig:patternmining}
\end{figure} 

\section{Expressivity of Rule Shapes}
\label{appx:ExpressivityOfRuleShapes}

Rule-learning approaches make trade-offs between the kind of rule structures that they are capable of mining and the scalability of the method. Below, we review the common rule classes, in decreasing generality, and then position SPECTRUM within this wider context.

\paragraph{Unrestricted ILP.} Systems such as ILASP \citep{law2020ilasp} and Popper \citep{cropper2021learning} support arbitrary Horn clauses along with recursion and predicate invention, but this added expressivity comes at the cost of much larger search spaces. This limits practical scalability. For example, ILASP-based approaches become intractable at datasets on the order of just $10$ objects once given an unconstrained hypothesis space, and while Popper scales to $10^3$--$10^4$ examples for simple lookup-style tasks, scalability degrades sharply once predicate invention or recursion is required \citep{pmlr-v162-glanois22a}.

\paragraph{Template-based rule mining.} Several learning systems, such as Metagol~\citep{cropper2016metagol}, search over user-provided metarules, rather than unrestricted clauses, to restrict the search space. In addition, differentiable ILP approaches, including $\delta$ILP~\citep{diffILP} and HRI~\citep{pmlr-v162-glanois22a}, construct rules by composing a fixed collection of protorules. In the case of HRI, the default protorules are:
\begin{enumerate}
    \item Chain: $Q(A, C) \land R(C, B) \rightarrow H(A, B)$
    \item Loop: $Q(A, B) \land R(B, A) \rightarrow H(A)$
    \item Reciprocal: $Q(A, B) \land R(B, A) \rightarrow H(A, B)$
\end{enumerate}

Repeated composition of these protorules allows increasingly expressive rule structures to be represented. For example, consider the rule
\[
\text{live}(X, Y) \land \text{lang}(Y, Z) \land \text{bornIn}(X, Y) \rightarrow \text{speak}(X, Z).
\]
This is not a chain rule, owing to the additional `branch' of conditional information, $\text{bornIn}(X, Y)$. However, this rule can be learned by HRI in two layers:
\[
\text{live}(X, Y) \land \text{bornIn}(X,Y) \rightarrow \text{aux}(X, Y) \quad \text{[reciprocal rule]}
\]
\[
\text{aux}(X, Y) \land \text{lang}(Y, Z) \rightarrow \text{speak}(X, Z) \quad \text{[chain rule]}
\]

Like HRI~\citep{pmlr-v162-glanois22a}, SPECTRUM can learn chain rules, loops, and reciprocals. However, unlike HRI, it cannot compose these to arbitrary depths. This results in a hypothesis space that is intermediate between chain-rule miners and miners such as HRI. However, HRI's generality comes at a scalability cost: HRI has been demonstrated to scale to graphs of up to $O(10^4)$ objects on tasks requiring predicate invention, but has not, to our knowledge, been evaluated beyond this scale. In addition, just because a rule is representable in the hypothesis space does not imply that it is easily findable by gradient descent. For this reason, HRI needs to inject noise into the learning process and execute repeated learning cycles to guarantee finding useful rules. By contrast, SPECTRUM does not have this limitation; all rules in its hypothesis space can be recovered, up to $\varepsilon$-uncertainty guarantees.

\paragraph{Chain rules.} A length $k$ chain rule can be expressed as:
\[
R_1(X_0, X_1) \land R_2(X_1, X_2) \land \dots \land R_k(X_{k-1}, X_k) \rightarrow H(X_0, X_k),
\]
where the variables form a simple, linear path. This is the most common case for scalable rule miners, such as AMIE~\citep{lajus_fast_2020}, Neural-LP~\cite{NeuralLP}, DRUM~\citep{DBLP:conf/nips/SadeghianADW19}, NCRL~\cite{cheng2023neural}, and RNNLogic~\citep{rnnlogic}.

\paragraph{Rules learnable by SPECTRUM.} Depending on the type of predicate in the head, SPECTRUM can find the following two shapes:
\begin{enumerate}
    \item \textbf{Binary head.} When the head predicate is binary, by construction, the learnt rule must form a closed path. In contrast to chain rules, the order of variables appearing in the binary predicates does not matter, and the rule can have up to one unary predicate per variable. For example,
    \[
    \text{friends}(X_1, X_2) \land \text{isVerified}(X_2) \land \text{friends}(X_2, X_3) \rightarrow \text{friends}(X_3, X_1)
    \]
    is a length-3 loop. All chain rules are a subset of this type.
    
    \item \textbf{Unary head.} When the head predicate is a unary predicate, the rule can form an open path, as long as the ends are capped by unary predicates. For example,
    \[
    \text{livesIn}(X, Y) \land \text{locatedIn}(Y, Z) \land \text{isDisasterZone}(Z) \rightarrow \text{isAtRisk}(X)
    \]
    is an open two-hop path ending in a unary.
\end{enumerate}

Two special cases outside this general pattern are the following:
\begin{enumerate}
    \item \textbf{Unary chain.} All atoms are unary and share the same single variable:
    \[
    U_1(X) \land U_2(X) \land \dots \land U_N(X) \rightarrow U_{N+1}(X).
    \]
    For example, $\text{isAdult}(X) \land \text{hasID}(X) \rightarrow \text{canVote}(X)$.
    
    \item \textbf{Binary rule on a fixed pair.} All atoms are binary and relate the same pair of variables (in any order):
    \[
    B_1(X, Y) \land B_2(Y, X) \land \dots \rightarrow B_{N+1}(X, Y).
    \]
    For example, $\text{follows}(X, Y) \land \text{follows}(Y, X) \rightarrow \text{mutualFollow}(X, Y)$.
\end{enumerate}




\end{document}